\documentclass[10pt]{article}
\usepackage[explicit]{titlesec}
\usepackage{hyphenat}
\usepackage{ragged2e}
\RaggedRight

\usepackage{rotating}  
\usepackage{theorem}   
\usepackage{bm}  
\usepackage{amsmath,amsfonts,amssymb}
\usepackage{booktabs}  
\usepackage{pdfsync}   
\usepackage{graphicx}
\usepackage{latexsym}
\usepackage{amsmath}
\usepackage{amssymb}
\usepackage{fancyvrb}
\usepackage{subfigure}
\usepackage{colortbl}
\usepackage{enumerate}
\usepackage{pifont}
\usepackage{stmaryrd}
\usepackage{textcomp}
\usepackage{fncylab}
\usepackage{multirow}
\usepackage{paralist}
\usepackage{wrapfig}
\usepackage{longtable}
\usepackage[]{algorithm2e}
\usepackage[table]{xcolor}
\usepackage{lscape}
\usepackage{lipsum}

\usepackage{garamondx}
\usepackage[T1]{fontenc}
\usepackage{amsmath}
\usepackage{graphicx}
\usepackage{xcolor}

\usepackage{geometry}
\usepackage{fancyhdr}
\usepackage[labelfont={footnotesize,bf} , textfont=footnotesize]{caption}
\makeatletter
\renewcommand\tagform@[1]{\maketag@@@ {\ignorespaces {\footnotesize{\textbf{Equation}}} #1.\unskip \@@italiccorr }}
\makeatother
\titleformat{\section}
  {\normalfont}{\thesection}{1em}{\MakeUppercase{\textbf{#1}}}
\titlespacing\section{0pt}{0pt}{-10pt}
\titleformat{\subsection}
  {\normalfont}{\thesubsection}{1em}{\textit{#1}}
\titlespacing\subsection{0pt}{0pt}{-8pt}

\makeatletter
\newcommand\sixteen{\@setfontsize\sixteen{17pt}{6}}
\renewcommand{\maketitle}{\bgroup\setlength{\parindent}{0pt}
\begin{flushleft}
\sixteen\bfseries \@title
\medskip
\end{flushleft}
\textit{\@author}
\egroup}
\makeatother

\usepackage[sort&compress]{natbib}
\makeatletter
\renewcommand\@biblabel[1]{\textbf{#1.}\hfill}
\makeatother

\bibpunct{}{}{,~}{s}{,}{,}
\title{An Intelligent Hybrid Model for Identity Document Classification}

\author{
Nouna Khandan*$^{a}$ \\ \medskip
$^{a}$Macquarie University, Sydney, Australia \\  \medskip
nouna.khandan@students.mq.edu.au
}

\begin{document}

\vspace*{.01 in}
\maketitle
\vspace{.12 in}

\section*{abstract}

Digitization, i.e., the process of converting information into a digital format, may provide various opportunities (e.g., increase in productivity, disaster recovery, and environmentally friendly solutions) and challenges for businesses. In this context, one of the main challenges would be to accurately classify numerous scanned documents uploaded every day by customers as usual business processes. For example, processes in banking (e.g., applying for loans) or the Government Registry of BDM (Births, Deaths, and Marriages) applications may involve uploading several documents such as a driver's license and passport. There are not many studies available to address the challenge as an application of image classification. Although some studies are available which used various methods, a more accurate model is still required. The current study has proposed a robust fusion model to define the type of identity documents accurately. The proposed approach is based on two different methods in which images are classified based on their visual features and text features. A novel model based on statistics and regression has been proposed to calculate the confidence level for the feature-based classifier. A fuzzy-mean fusion model has been proposed to combine the classifier results based on their confidence score. The proposed approach has been implemented using Python and experimentally validated on synthetic and real-world datasets. The performance of the proposed model is evaluated using the Receiver Operating Characteristic (ROC) curve analysis.

\section*{keywords}
Business Process Management; Image Processing; Document Classification

\vspace{.12 in}


\section{Introduction}

Business processes are central to the operation of any organization~\cite{DBLP:journals/corr/abs-2105-10911,DBLP:journals/spe/BeheshtiBM18,DBLP:books/sp/BeheshtiBSGMBGR16,DBLP:conf/bpm/BeheshtiBNS11}.
Today, digital technology is integrated into all aspects of business processes focusing on improving the quality of services~\cite{DBLP:conf/birthday/YangTB21,DBLP:conf/birthday/BarukhZBBBCYSS21,DBLP:conf/edoc/JalayerKBPM20,DBLP:conf/wise/BeheshtiBSS19}.
Several organizations and governments have carried out digital transformation projects, e.g., an automotive manufacturer increasing traffic safety based on big data~\cite{DBLP:journals/dpd/BeheshtiBTMBN19,DBLP:conf/wise/BeheshtiBSS19}. The goal is to improve digital services and replace manual processes~\cite{DBLP:conf/assri/ShahbazBNQPM18,DBLP:conf/bpm/BeheshtiSGABYSC18,DBLP:conf/icsoc/SchiliroBGABYSC18,DBLP:conf/icsoc/AmouzgarBGBYS18}.
For example, the UK government in 2012 started a trusted online communications and public services channel which provides various digital services and supported 14 million visitors per week in 2019. In Australia, myGov\footnote{https://my.gov.au/mygov/content/html/about.html} is the secure way to access the government's digital services, such as job search, taxation office, and child support. The three strategic priorities for the digital transformation of the Australian government are: easy to deal with, driven by users, and fit for every user. This clearly shows the importance of user experience and that making the services delighter is an essential key to success. To identify and authenticate the people who access online government services, the Australian government is now working on GovPass\footnote{https://www.dta.gov.au/our-projects/digital-identity/trusted-digital-identity-framework}  which is their newest digital identity solution.
%

To use the mentioned online digital services, people need to prove their identity which is confidential and crucial. Moreover, governments and businesses also require to identify the users to detect and avoid fraudulent activities. Authentication of the identity documents in terms of information accuracy and originality is a way to achieve the goal. myGovID\footnote{https://www.mygovid.gov.au/} is the Australian governmental service in which users scan one of their identity documents for authentication. The scanned document is then checked against all existing government records. As users can provide any of their ID documents, the application searches through all ID document databases to find the related record. However, searching through all types of databases takes time. Defining the type of the scanned document can enhance the accuracy of the possible OCR (Optical Character Recognition) method and reduce search time by searching in only one specific database.

Furthermore, many other services have been digitized in which uploading several documents are required to enable clients to use that specific service. For instance, processes in banking (e.g., applying for loans) or the Government Registry of BDM (Births, Deaths, and Marriages) are two examples of services that demand uploading several documents, as well as manually providing the details of the uploaded documents, which is very error-prone and frustrating. Automatically defining the type of the documents enables the mentioned type of services to eliminate the need for the user to enter relevant information, y automatically extracting the information from the specified document. In this context, one of the main challenges would be to accurately classify a huge number of scanned documents uploaded every day by customers as part of organizations' business processes. The classification of these documents, which are mostly low-quality images, affected by different lighting conditions, perspectives, and rotation, is challenging and demands a robust classification model.

\subsection{Problem Statement}

Digitization, i.e., converting information into a digital format, may provide various opportunities (e.g., increase in productivity, disaster recovery, and environmentally friendly solutions) and challenges for businesses. In this context, one of the main challenges would be to accurately classify a large number of scanned documents uploaded every day by customers as part of organizations' business processes. For example, processes in banking (e.g., applying for loans) or the Government Registry of BDM (Births, Deaths, and Marriages) applications may involve uploading several documents such as a driver's license and passport. There are not many studies available to address the challenge as an application of image classification. Although some studies are available which used various methods, a more accurate model is still required.

\subsection{Contribution}

This study proposes an intelligent hybrid model based on different machine learning hybrid models to classify Identity Documents. First, a fundamental paper in which the dataset and results meet our project conditions has been replicated. By analyzing the results of the replicated model and the characteristics of the current study dataset, the model has been improved by adjusting the methods based on the existing imagery dataset. The final proposed model that results in 100\% accuracy is based on two different methods. First, the SIFT (Scale-Invariant Feature Transform) feature extractor, with which the image is classified based on its visual features. Second, OCR (Optical Character Recognition), with which the identity documents are classified based on their text features. To combine the two mentioned classifiers, a novel model based on statistics and regression has been proposed to calculate the confidence levels of the SIFT classification results. By applying a fuzzy mean approach on the confidence levels from each classifier, the overall classification result has been calculated. Finally, a crowdsource technique has been proposed as future works, in which the samples with the lower classification confidence level derived from the fusion model will be passed to a crowdsourcing system. The mentioned documents will be shared with crowd workers, and the crowd's knowledge will be used to facilitate a systematic arrangement of classes' items in categories according to established criteria shared by the crowd workers. A feedback loop has been presented to continuously take samples from the classes and refine the categories of classified documents.


In this section the generalities of the research are introduced, including the explanation of the main issue discussed in the research, the importance and necessity of the research, the goals and methodology of the research. The rest of the present research report is organized as follows:
\textbf{Section 2} reviews the technical background of the current study and the specific technique used for the accuracy enhancement of the results of similar projects. Also, section 2 contains the literature on the subject and the contents of previous research done (or ongoing) related to the research subject.
\textbf{Section 3} which includes the core of the research, introduces the proposed approach for defining the type of identity documents based on their images. In this section all steps that have been taken to improve the accuracy of the final proposed model, have been explained. Also, a brief explanation of the preliminary method implementation, as well as the advantages and disadvantages of each model, have been analyzed.
\textbf{Section 4} contains the validation of the method, which was done through its implementation and software testing. We describe the software, packages, libraries, and methods used to implement the final proposed model. Then we demonstrated the results of the model using different validation methods.
\textbf{Section 5} concludes the study by review and comparison of the results of different methods that we have taken, as well as a brief comparison and evaluation of the results of the current study with previous studies which have been analyzed in section 2. finally, the future work of the research is summarized and introduced.


\section{Background and State-of-the-Art}
This section provides a brief overview of previous or ongoing activities and research related to the research topic. The prior researches, which form the current research's information infrastructure, is examined step by step in the continuation of the report, and then, in terms of the new idea presented in this research, they are thematically expanded. By considering the multidimensionality of the information infrastructure of the research subject, the materials related to the research background have been prepared and presented in different titles and dimensions.
In this section, the focus has been reviewing the research done in recent years to receive and document the subject's latest scientific situation and achievements. Research from previous and older years has also been reviewed if it contains essential scientific material. This section includes three sections; the first section demonstrates the background of image processing. The second section analyzes the state of the art in ID document classification. Finally, the third section discusses crowdsourcing as a common post-processing model to improve classification model results.

\subsection{Image Classification and Recognition}

Image classification and recognition is an actively pursued area in artificial intelligence and computer vision. It has gained ground to such an extent due to two important reasons: The potential to replace human vision. Such empowered machines are capable of reducing the workload and errors of humans in mission-critical industries such as medical or defense \cite{javidi2002image}. Image Recognition gives machines the ability to understand and analyze images to automate a specific task, such as image classification, detection, and segmentation.

This section focuses on the current state-of-the-art and relevant concepts in Image Classification. At first, the focus is on the image classification process and then will investigate different classification approaches and techniques.
Although identifying objects in images and classifying them are easy tasks for a human being, they have been complex problems for machines. With recent accelerating advances in producing high-capacity computers and optoelectronic devices, image recognition and classification systems have become one of the most popular academic research areas in Image Processing. A plethora of research projects has been carried out to facilitate image classification in different categories.
Image classification approaches can be categorized into five different types, which have been investigated widely in the literature \cite{nath2014survey, lu2007survey}.

\subsubsection{Image Classification Process}

\begin{figure}[h]
    \centering
    \includegraphics[width=1.0\textwidth]{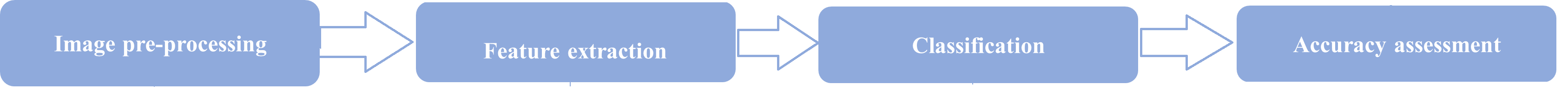}
    \caption{Image classification process.}
    \label{fig:02-02}
\end{figure}

There are several surveys to distinguish which approach is better for a specific task; however, regardless of the method used, they follow the four major steps~\cite{nath2014survey,lu2007survey} of image pre-processing, feature extraction, classification, and accuracy assessment which is explained below and shown in Figure~\ref{fig:02-02}. Many researchers prefer to focus on one single step and make innovation improvement ideas \cite{li2018image, wang2015use}.
\begin{itemize}
    \item\textbf{image pre-processing } To improve the raw images, which may have dramatic positive enhancements on the quality of feature extraction and the result of image classification. The steps to be taken are: reading an image, resizing the image, denoising, and smoothing edges \cite{krig2016image}.
     \item\textbf{Feature extraction} After the pre-processing, to identify helpful features in classification, feature extraction techniques will be applied, and relevant information from the input image data for class characterization will be extracted \cite{li2018image, kumar2014detailed}.
    \item\textbf{Suitable classification approach selection} A suitable classification method to categorize detected objects into predefined classes to be selected. Major classification approaches and techniques will be presented in sections 1.2 and 1.3.
    \item\textbf{Validation} Validation or accuracy assessment is the final step in the image classification process. It is carried out once the classification has been completed and is defined as an integrated part of any classification project. Different projects have different accuracy requirements, and the classifiers that show a low level of accuracy will be crossed out \cite{anand2017block}.
\end{itemize}

\textbf{A - Image Pre-Processing}
In this section, the problems of image quality, corrections, and enhancements have been investigated. Digital devices may produce raw image data with different kinds of issues. The quality of the image data can be improved by removing distortions or enhancing some image features. These improvements play an important role in the next steps of image processing \cite{krig2016image}.
According to the size of the pixel neighborhood, four categories of image processing are defined: pixel brightness transformations, geometric transformations, local neighborhood of the processed pixel-based method, and image restoration \cite{sonka1993image}.

\begin{figure}[h!]
    \centering
    \includegraphics[width=0.5\textwidth]{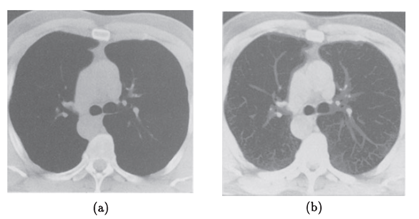}
    \caption{Pixel brightness transformation. (a) original image, (b) modified image \cite{sonka1993image}.}
    \label{fig:02-03}
\end{figure}

\begin{figure}[h]
    \centering
    \includegraphics[width=0.5\textwidth]{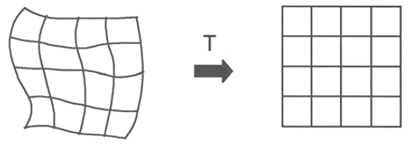}
    \caption{Geometric transform on a plane \cite{sonka1993image}.}
    \label{fig:02-04}
\end{figure}

\begin{figure}[h!]
        \centering
        \includegraphics[width=0.7\textwidth]{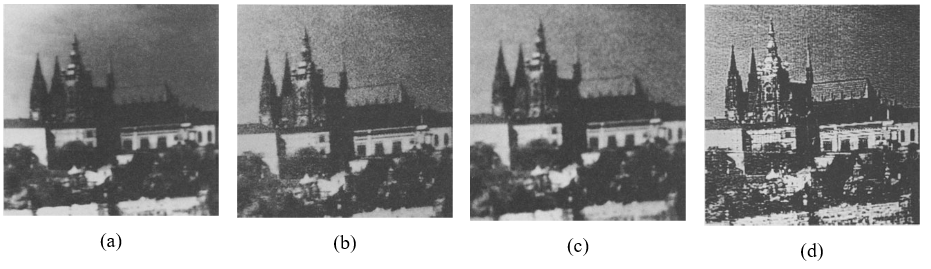}
        \caption{Local Pre-processing techniques. (a) Original Image, (b) Superimposed Noise, (c) Smoothing the image, (d) Gradient Operator over the image Restoration of motion blur image \cite{sonka1993image}.}
        \label{fig:02-05}
    \end{figure}

\begin{itemize}
    \item\textbf{Pixel brightness transformations}\\
    There are two classes of pixel brightness transformations: Brightness corrections, which enable modification of the brightness of the pixel by considering its position and original brightness. Figure \ref{fig:02-03} depicts the effect of brightness transformation over an image. And greyscale transformations adjust brightness without considering its position in the image.
\end{itemize}
\begin{itemize}
    \item\textbf{Geometric transformations}\\
    Geometric transformations are widely used in computer graphics and image analysis. It can eliminate the geometric distortions that may occur when an image is taken (Figure \ref{fig:02-04}). Some important geometric transformations are:
    (i)~Rotation; (ii)~Change of scale; and (iii)~Skewing by the angle.
    Article \cite{lucas2016image} as a geometric transformation approach proposes a new approach to allow digital watermark detection and extraction.
\end{itemize}
\begin{itemize}
    \item\textbf{Local Pre-Processing}\\
    Two groups of local pre-processing methods depending on the goal of the processing can be defined: (1) Image smoothing and (2) Gradient operators.
    \begin{itemize}
        \item\textbf{Image smoothing} aims at suppressing image noise. It can remove impulsive noise or thin strips due to degradation. Smoothing also can blur sharp edges that destroy the image. In this case, image restoration techniques can be used. Image restoration will be discussed in section D of the pre-processing step \cite{sonka1993image}.
        \item\textbf{Gradient operators} aim at finding locations in the image where there are rapid changes. It has an effect like Fourier transform domain in suppressing low-frequency noises.
    \end{itemize}

As a drawback, the noise level increases after applying gradient operators on an image. Many surveys are addressing this problem \cite{sonka1993image}. For example, Senel et al.\cite{senel2007topological} claim that by applying its method to synthetic and natural images.
Smoothing and gradient operators have conflicting aims. Rabiee et al. \cite{rabiee1993new} propose an algorithm to solve this problem by allowing concurrent smoothing and edge enhancement through a robust filter. Figure \ref{fig:02-05} illustrates the effect of smoothing and gradient operators over an image.

\end{itemize}
\begin{itemize}
    \item\textbf{Image restoration}\\
A pre-processing method that suppresses the degradation of an image based on the knowledge of its nature and is mostly based on convolution applied to the whole image. Sub-optimal lenses, nonlinear electro-optical sensors, motion, wrong focus, etc., are examples of causes of image degradation. Figure \ref{equ:02-06} is a restoration of motion blur example.
    \begin{figure}[h!]
        \centering
        \includegraphics[width=0.45\textwidth]{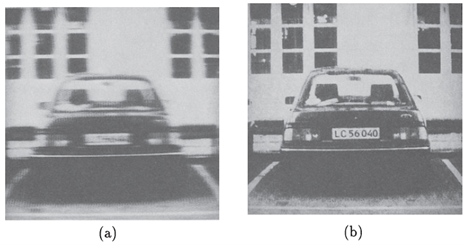}
        \caption{Restoration of motion blur image \cite{sonka1993image}.}
        \label{fig:02-06}
    \end{figure}

Two groups of Image restoration techniques have been described below: In ~\textbf{deterministic methods}, an image with known degradation function and little noise can be restored by applying a transformation inverse to the degraded image; and in~\textbf{Stochastic methods} the best restoration function by applying a particular stochastic measure can be found.
Medjahed et al. \cite{medjahed2015comparative} combine the above-mentioned methods with the objective of enhanced image reconstruction with improved resolution performances. 
\end{itemize}

\textbf{B - Feature extraction}
After the pre-processing phase, feature extraction will be done as an essential step in image processing. It allows a perfect representation of the image content. Feature Extraction mainly aims at extracting relevant information from an image that characterizes classes. In this process, feature vectors will be generated based on relevant features then will be used by classifiers to identify the input unit with a specific output unit \cite{kumar2014detailed}. To differentiate various classes, the extracted features must be informative, the features should avoid the irrelevant knowledge in the input, also, the number of the features should be limited \cite{lippmann1989pattern}.\\

\textbf{B.1 - Feature Extraction Techniques \cite{medjahed2015comparative}}\\

There are different feature extraction techniques available. Color features, texture features, and shape features are the three most important features which have been touched on in this section.
\begin{itemize}
    \item\textbf{Color features} The color is the most important and the simplest feature that humans recognize when observing an image. The color histogram is a commonly used method to extract color features. It is based on the distribution of the color in an image but not the size, rotation, or zoom of the image. \\
    Several internationally well-known color spaces exist that their color spaces were designed for a particular use, such as RGB color space, Munsell color system, CIE color systems, and HSV color space. The aim is to identify a feature to find the similarity between the features of two images \cite{kavitha2016texture}.
    \item\textbf{Texture feature} A group of pixels that has certain characteristics is called texture. Some describe it as a coarseness, contrast, line-likeness, regularity, and roughness measure. Others described it as repetitive regions of pixels in an image. Regardless of the definitions, the texture feature approaches are classified into seven categories: statistical, structural, transform-based, model-based, graph-based, learning-based, and entropy-based \cite{humeau2019texture,DBLP:conf/adc/RastanPSRB18,abu2021relational}.\\
    \item\textbf{Shape features} or object recognition or shape description are to find all translated, rotated, and scaled instances of a given image within a database.
    The shape feature extraction methods are contour-based and region-based. (1) The contour methods focus on the boundary. (2) the region methods consider the entire region \cite{medjahed2015comparative, saber1996integration}.\\
\end{itemize}

\textbf{B.2 - Feature Extraction Methods \cite{medjahed2015comparative}}\\

A Portion of feature extraction methods proposed in the literature are an integration of the above-mentioned techniques \cite{kavitha2016texture, ganar2014enhancement, saber1996integration}.
Three popular methods with wide application and good performance are  Scale Invariant Feature Transformation (SIFT) \cite{lowe2004distinctive}, Speeded-Up Robust Features (SURF) \cite{bay2006surf} and Histogram of Oriented Gradient (HOG)~\cite{dalal2005histograms}.

\begin{itemize}
\item \textbf{Scale Invariant Feature Transformation (SIFT)}\\
SIFT is a texture feature extraction technique with the robust matching between various views of an image. It consists of three main computational stages: Scale-space extrema detection, orientation computation, and keypoint descriptor \cite{lowe2004distinctive}.

\begin{itemize}
    \item\textbf{Scale-Space Extrema Detection}\\
    The first step is to find the keypoints. This method uses the Gaussian kernel, which is the best kernel for scale-space in images. In this stage, it uses the Difference of Gaussians (DoG) function to identify keypoints. To improve the computational speed, instead of Gaussian,  DoG scale space is obtained by subtracting adjacent surfaces from each other \cite{juan2009comparison}.

\end{itemize}
\begin{itemize}
    \item\textbf{Orientation Computation}\\
    In this step, the position of each keypoint is precisely determined in terms of coordinates, and the keypoints that have low contrast will be eliminated. Then based on the image gradient at each keypoint location,  the orientations are assigned. 

\end{itemize}
\begin{itemize}
    \item\textbf{Keypoint descriptor}\\
     The computation of the keypoint descriptor is shown in Figure \ref{fig:02-08}. First, around each keypoint, gradient orientations are measured. Then, based on the keypoint orientation, descriptor orientations are rotated to achieve orientation invariance \cite{yang2008image}.
    For efficiency, the gradients are calculated for all the levels of the pyramid as it is mentioned in the previous phase. They are depicted with small arrows on the left side of Figure \ref{fig:02-08}. A Gaussian weighting function is considered to assign a weight to the magnitude of each sample point. This is showed with a circle window on the left side of Figure \ref{fig:02-08}. Then, as it is illustrated on the right side of Figure \ref{fig:02-08}, the samples are accumulated into orientation histograms.\\

    \begin{figure}[h]
    \centering
    \includegraphics[width=0.4\textwidth]{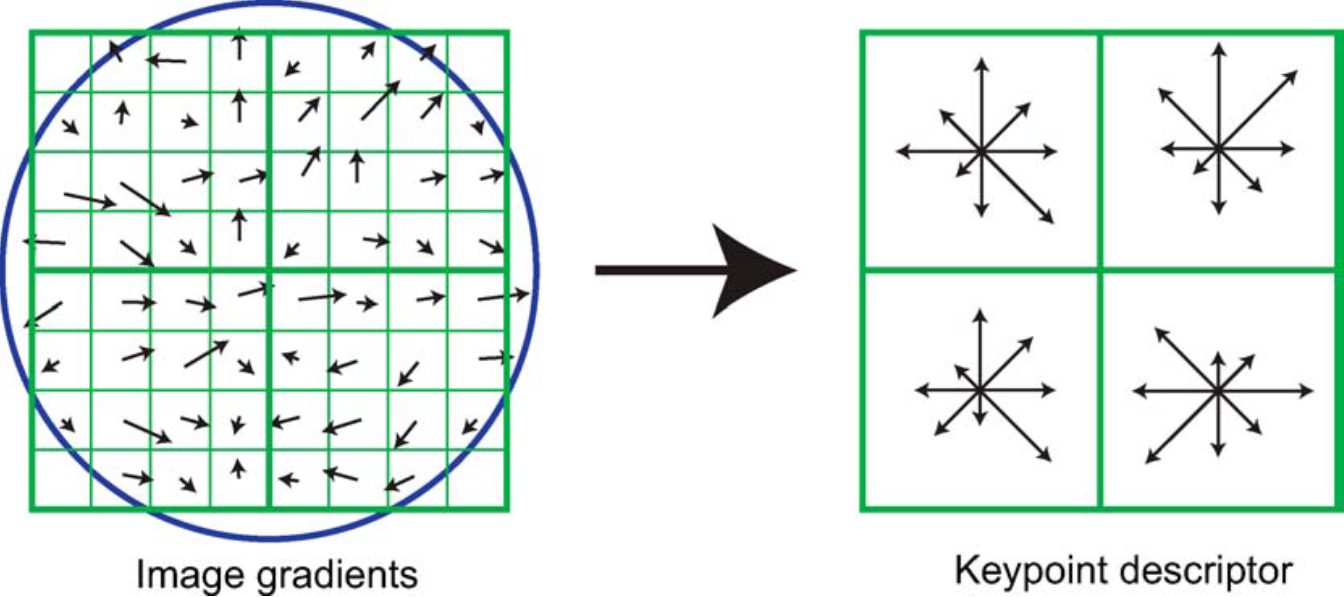}
    \caption{The computation of keypoint descriptors from the image gradients \cite{lowe2004distinctive}.}
    \label{fig:02-08}
    \end{figure}
\end{itemize}
\end{itemize}
\newpage
\begin{itemize}
\item \textbf{Speeded-Up Robust Features (SURF) \cite{bay2006surf}}\\
SURF is a scale and rotation-invariant point detector and descriptor based on integral images. It is motivated by the SIFT algorithm with two similar stages: keypoint detection and description.

\begin{itemize}
    \item\textbf{Keypoint detection}\\
    The first step here is also finding the keypoints. But instead of using DoG, which is used in SIFT, SURF uses integral images \cite{yang2008image}.
    The integral image, \textit{I(X)}, at a location like \textit{X = (x,y)} is the sum of all pixels in the input image \textit{I(i,j)} within a rectangular region.

    \begin{equation}\label{equ:02-06}
    I(X)= \sum_{i=0}^{x}\sum_{j=0}^{y}I(i,j)
    \end{equation}

    Due to using the integral images, it only takes four additions to calculate the sum and requires a smaller number of operations to complete the convolution. So, the speed of the process is improved over the SIFT algorithm. Here, the keypoint detector is based on the Hessian matrix because of its fast computation and high accuracy. By having the location of a point \textit{X = (x,y)} in an image I, the Hessian matrix \textit{H(x, $\sigma$)} in \textit{X} at scale \textit{$\sigma$} is defined as follows:

    \begin{equation}\label{equ:02-07}
    H(X,\sigma)=\begin{bmatrix} \L_{xx}(X,\sigma) & \L_{xx}(X,\sigma)\\ \L_{xx}(X,\sigma) & \L_{xx}(X,\sigma) \end{bmatrix}
    \end{equation}

\end{itemize}
\begin{itemize}
    \item\textbf{Keypoint description}\\
    To extract the SURF descriptor, two steps are taken: First, fixing a reproducible orientation based on information from a circular region around the interest point. Second, constructing a square region aligned to the selected orientation and extract the SURF descriptor from it. In order to assign orientation, Calculated responses in x and y directions shown in Figure \ref{fig:02-02}, in the page \pageref{fig:02-02}.
\end{itemize}
\end{itemize}
\begin{itemize}
\item \textbf{Histogram of Oriented Gradient (HOG) \cite{dalal2005histograms}}\\
Hog detects the shape structure of objects in images and is robust to illumination change and scaling. So, it is widely used in object detection \cite{saber1996integration}. Hog characterizes local object appearance and shape by the distribution of local intensity gradients or edge location.
HOG feature extraction mainly includes two steps: Firstly, extracting the magnitude and direction of gradient from the raw image to generate the gradient magnitude histogram.

Secondly, since the HOG descriptor is based on image gradient, the whole image is divided into small sub-images called a cell. Within each cell, the gradient histogram of all pixels is calculated in terms of the direction bin. And then, the histograms build up one vector, which represents the desired descriptor \cite{saber1996integration, ge2017texture}.
HOG has two main defects: being sensitive to rotation transformation and having a large feature dimension which increases the computation cost in training. Hence, many efforts have been taken to improve the model. Ge et al. \cite{ge2017texture} have proposed Histogram of Oriented Gradient Domain Texture Tendency (HOGTT) algorithm claiming that they can make up for the above-mentioned defect of HOG by developing an efficient texture tendency-based Hog feature that is invariant to illumination, scaling, and rotation.
\end{itemize}

\textbf{C - Suitable classification approach selection}
At this point, by using a suitable classification technique, detected objects will be categorized into predefined classes.
There are five different methods. The classification techniques will be discussed separately in sections 2.1.2 and 2.1.3.

\textbf{D - Validation}
The final step in the image classification process is validation. The goal here is to verify how effectively the pixels of an image are categorized into the correct classes. It compares the classified image to another source of data that is considered to be accurate data \cite{lu2007survey}.

\textbf{Confusion Matrix}\\
The confusion matrix is a commonly used method that quantifies the accuracy of image classification. It is in the form of a table that displays the correspondence between reference image data and the classification result. Figure \ref{fig:02-T01} is a sample confusion matrix.
In this table, the number of classified images is assumed three (a, b, and c), and there are three reference classes like A, B, and C. The columns are truth classes, and the rows are the classified images that should be assessed \cite{anand2017block}. The overall accuracy can be calculated based on these cells per Equation \ref{equ:02-10}:

    \begin{figure}[h!]
    \centering
    \includegraphics[width=0.6\textwidth]{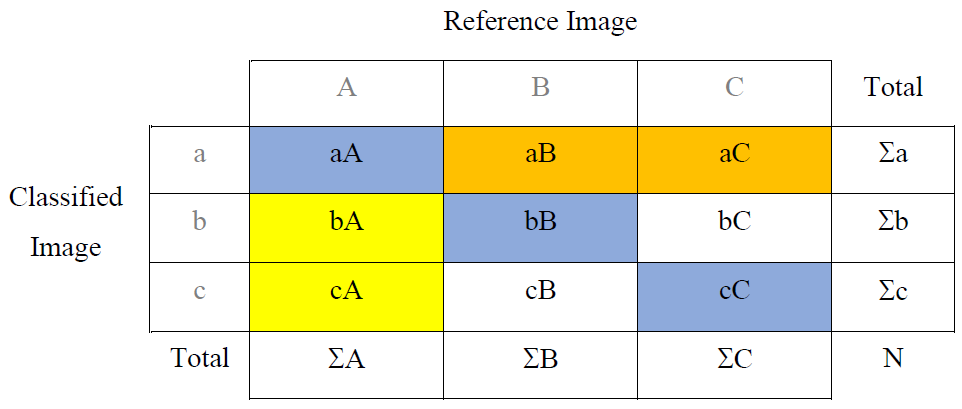}
    \caption{A sample confusion matrix.}
    \label{fig:02-T01}
    \end{figure}


\begin{equation}\label{equ:02-10}
    Overall\; accuracy=\frac{(aA+bB+cC)}{N}
\end{equation}

\subsubsection{Classification Approaches}

\textbf{A - Based on nature of training sample used in classification}
\begin{itemize}
    \item\textbf{Supervised classification} refers to a process in which the classification is supervised. Prior knowledge is essential before the testing phase. Nath et al. \cite{nath2014survey} suggested four steps in this process: (1) Identifying the training areas for each informational class, (2) Signatures identify, (3) Classifying all pixels, (4) Mapping of the information class. Maximum likelihood, minimum distance, artificial neural network (ANN), and decision tree are examples of supervised classification approaches \cite{carneiro2007supervised, baxi2011supervised}.
    \item\textbf{Unsupervised classification} refers to a process in which a large number of unknown pixels will be examined and divided into a number of classes based on their nature. No prior knowledge or human intervention is required for this process \cite{lu2007survey}. Nath et al. \cite{nath2014survey} proposed four steps in this technique: (1) Clustering the data, (2) All pixels are then classified based on clusters, (3) spectral class-map, (4) cluster labeling done by the analyst, (5) map the informational class. K-means clustering algorithm is an example of an unsupervised classification approach \cite{prajapati2015supervised, hellmann1998new,niu2020ensemble}.
\end{itemize}

\textbf{B - Based on the basis of various parameter used on data}
\begin{itemize}
    \item\textbf{Parametric classifiers:} In a parametric classifier, the statistical probability distribution of each class is based on classification.
    Bayesian, Multivariate Gaussian, Linear discriminate analysis, and maximum likelihood are examples of this approach \cite{lu2011parametric, cobra1994classification}.
    \item\textbf{Non-parametric classifiers:} Non-parametric classifier is used when an unknown density function is used to estimate the probability density function. 
    Artificial Neural Networks (ANN), Support Vector Machine (SVM), Decision Tree classifier, and expert system are examples of non-parametric classifiers \cite{wang2015use, chen2012svm}.
\end{itemize}

\textbf{C - Based on the nature of pixel information used on data}
\begin{itemize}
    \item\textbf{Per-Pixel Classifiers:} By considering the spectral similarities of a pixel with classes \cite{putri2019comparing, zoleikani2017comparison} it will be assigned to a class based on either parametric or non-parametric. 
    Neural networks, SVM, and decision trees are examples of suitable techniques to enhance classifications \cite{lu2007survey} based on the per-pixel method.
    \item\textbf{Sub-Pixel Classifiers:} In this method, a pixel will be assigned to various classes based on to the area occupied by that class within the pixel \cite{kumar2007subpixel}. Handling mixed pixel problem and information loss prevention make it suitable for low spatial resolution images \cite{putri2019comparing}. Spectral mixture analysis, subpixel, and Fuzzy-set classifiers are popular sub-pixel classifiers.
    \item\textbf{Object-Oriented Classifiers:} The object-based classifier not only considers the spectral values stored in digital number (DN) but also counts on topologic (as neighborhood, contextual) and geometric (as size, shape) as classification parameters \cite{putri2019comparing, zoleikani2017comparison,DBLP:conf/momm/BeheshtiHY19}. E-Cognition is an example of this type classifier. 
    \item\textbf{Per-Field Classifiers:} This classifier is specifically designed for handling environmental heterogeneity problems and also improving the accuracy of classification \cite{smith2001integrated}. The Geographic Information System or GIS-based classification technique is an example where vector data is used for image parceling and classification \cite{lu2007survey}.
\end{itemize}

\textbf{D - Based Upon The Number Of Outputs Generated For Each Spatial Data Element}
\begin{itemize}
    \item\textbf{Hard Classification:} Also called crisp classification. Each pixel is considered a unique class. The Maximum likelihood, Minimum distance, ANN, decision tree, and SVM are examples of hard classification.
    \item\textbf{Soft classification:} Also known as Fuzzy classification, pixels are allowed to belong to numerous classes. A Fuzzy-set classifier is an example of soft classification.
\end{itemize}

\textbf{E - Based Upon The Nature Of Spatial Information}
\begin{itemize}
    \item\textbf{Spectral Classifiers:} Pure spectral information will; be used in this method. The Maximum likelihood, Minimum distance, ANN are some examples of Spectral classifiers.
    \item\textbf{Contextual Classifiers:} Spatially neighboring pixel information. Point-to-point contextual, correction and  frequency-based are examples of contextual classifiers.
    \item\textbf{Spectral-Contextual Classifiers:} Both spectral and spatial information are used in this method. A combination of parametric or non-parametric and contextual algorithms are examples of this method.
\end{itemize}

\subsubsection{Advanced Classification Technique Samples}

\textbf{A - Artificial Neural Networks (ANN)}
ANN is a supervised non-parametric classifier. It is inspired by the human nervous system. So, it attempts to simulate the network of neurons of a human brain and make decisions in a human-like manner. Biological Neural Network and ANN are similar in structure. The Multi-Layer Perceptron (MLP) is the most basic model that simulates the transmission function of the human brain. Each of the neural cells of the human brain, known as a neuron, undergo processing after receiving input and transmits the result to another cell. This behavior continues until a certain result. In ANN, the first is the training mode, where the neuron learns to perform an operation. Then in the using mode, the learned information will be used on a sensed input to predict the output \cite{burges1998tutorial}. The output of a MLP-ANN can be calculated using Equation \ref{equ:02-17}.

    \begin{equation}\label{equ:02-17}
    Out=f(\sum_{i=1}^{n}x_iw_i+b)
    \end{equation}

\begin{figure}[h!]
    \centering
    \includegraphics[width=0.4\textwidth]{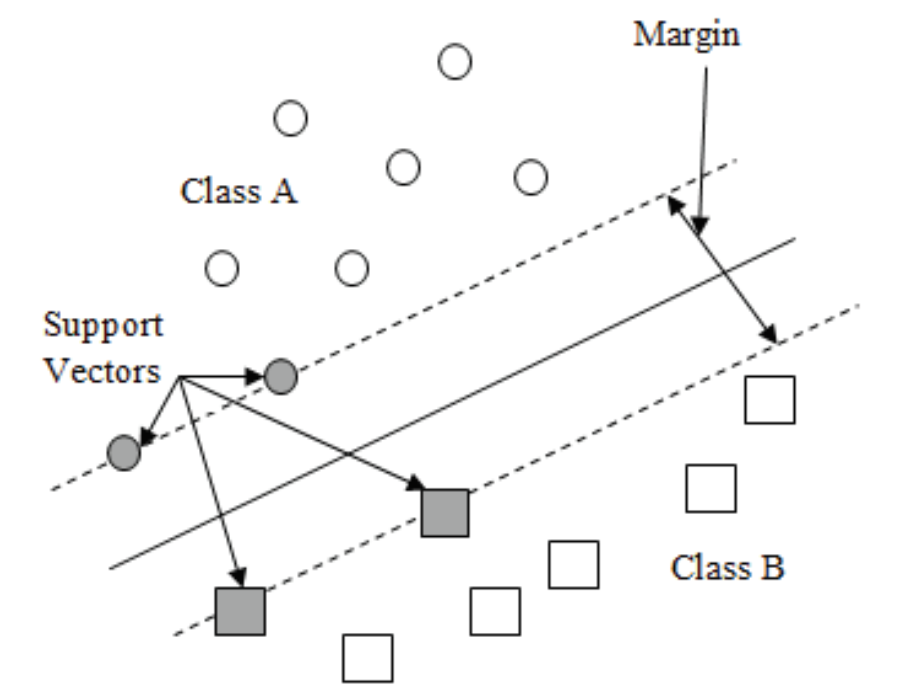}
    \caption{Separating classes A and B using SVM \cite{beaula2016comparative}.}
    \label{fig:02-11}
    \end{figure}

\textbf{B - Support Vector Machine (SVM)}
SVM is also a supervised non-parametric classifier. It is a binary and multi-class classifier that separates the classes by using a linear boundary.  SVM  tries to find a hyperplane in an N-dimensional space and perform a classification over them by maximizing the margin between data points (Figure \ref{fig:02-11})~\cite{beaula2016comparative}.

To calculate the gradients, the partial derivatives of the loss function will be considered. After simplifying, we have:

    \begin{equation}\label{equ:02-18}
    =2\lambda w_{k}+\left\{\begin{matrix} 0\;\;\;\;\;\;\; if\;y_{i}\left \langle x_{i},w \right \rangle\geqslant1 \\ -y_{i}x_{ik}\;\;\;\;\;\; else \end{matrix}\right.
    \end{equation}

Gradient update is as follows:
\begin{itemize}
\item When the model correctly predicts the classes:
    \begin{equation}\label{equ:02-19}
    w=w-\alpha .(2\lambda w)
    \end{equation}
\item When the model correctly predicts the classes:
    \begin{equation}\label{equ:02-20}
    w=w+\alpha .(y_{i}x_{i}-2\lambda w)
    \end{equation}
\end{itemize}

The performance and accuracy of this method are based on the hyperplane selection and kernel parameter. The linear , polynomial , and Gaussian kernels are examples of popular Kernel functions.

\begin{figure}[h!]
    \centering
    \includegraphics[width=0.5\textwidth]{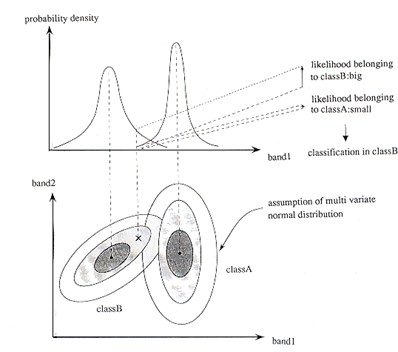}
    \caption{Concept of maximum likelihood method \cite{burges1998tutorial}.}
    \label{fig:02-12}
    \end{figure}

\textbf{C - Maximum Likelihood Classification (MLC)}
MLC is a supervised parametric classifier that takes the probability value of pixels into consideration to classify the pixels. Figure \ref{fig:02-12} illustrates the concept of MLC. In this method, by comparing the calculated probability of each pixel belonging to a class,  the pixels will be assigned to classes with the highest likelihood value. In the case of normal distribution, the likelihood function is \cite{al2009comparison, burges1998tutorial}:

\begin{equation}\label{equ:02-21}
    L_{k}(x)=\frac{1}{(2\pi)^{\frac{n}{2}}\left | \sum_{k}^{}\right |}\exp (-\frac{1}{2}(x-\mu _{k})^{T}\sum_{k}^{-1}(x-\mu _{k}))
\end{equation}

\textbf{D - Fuzzy Classifier}
Classification and pattern recognition are based on the disjoint of categories. In other words, no observation can belong to more than one class at the same time. As mentioned before in section 1.2 about Hard Classification, the algorithms that create such labels are called crisp labeling. In contrast, fuzzy sets assign more than one class to each observation and define a membership degree, exactly like fuzzy logic, which assigns a degree of correctness to the accuracy of a statement. Hence, such algorithms are called fuzzy classifiers and the resulting labels as soft labeling. In this way, each observation with different degrees or values can belong to more than one class or group \cite{mendel1995fuzzy}.\\
According to this, we can consider the fuzzy classifier D as an Approximator Function based on a set of features to create labels with weights in [0,1] intervals. We denote this proposition as follows using a mathematical notation~\cite{ludmila2008fuzzy}.

\begin{equation}\label{equ:02-22}
D: F \to [0,1]^c
\end{equation}

\begin{figure}[h!]
    \centering
    \includegraphics[width=0.5\textwidth]{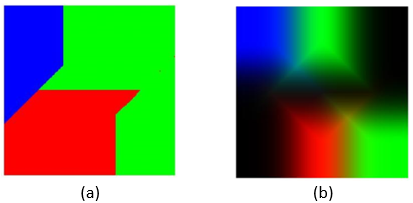}
    \caption{(a) Hard classification, (b) Soft classification \cite{ludmila2008fuzzy}. }
    \label{fig:02-13}
    \end{figure}

Where F is the feature space, and \textit{c} is the number of classes. \\
As a result, the implementation of fuzzy classification algorithms for each observation is a vector that has \textit{c} rows, and its elements are in [0,1] intervals. Thus, the value of the element \textit{i} of this vector determines the \textit{i-th} observation's membership degree of \textit{i-th} class. While creating such estimator functions is difficult by using other classification algorithms, fuzzy classifiers work well.
As shown in Figure \ref{fig:02-13}, three classes are taken into account; class "1" is encoded as red,  "2" as green, and "3" as blue. In Figure (a), the areas are formed using a hard classification with restricted boundaries. In Figure (b), there is a dark region between colors. The dark area shows the points where low membership values in three classes are identified~\cite{ludmila2008fuzzy}.

\subsection{ID Document Classification}
There are limited researches proposing different solutions for image based classification of the ID documents,  A summary of the most relevant ones have been reviewed and analyzed, including the reported results, can be found in this section.

Simon et al.~\cite{simon2015fine} proposed a method of identifying the various forms of identities using visual approaches. Their objective was to recognize the issuing country, document form, and document version. Although correct identification of pieces is easy but consistent recognition of a state is complex due to the wide variety of paper forms. Therefore, they used multiple approaches for this application based on recent advances using convolutional neural networks~\cite{DBLP:conf/intellisys/SchiliroBM20,DBLP:conf/ijcnn/KhatamiNB0NZ20,DBLP:journals/cee/NiuXAPBA20,DBLP:conf/wsdm/BeheshtiHYMG020}. They claimed that for specific scenarios, general-purpose OCR failed, so data classification was done first to make data recovery easier. The authors employed an open dataset of 375 differentially categorized documents into 74 classes. Using a single training picture, the combination of HOG and Color-name hit mean class-wise accuracy of 97.7\%. The authors found out that unknown records are often characterized by logistic regression and have relatively good accuracy of 0.986~\cite{simon2015fine}.

Awal et al.~\cite{awal2017complex} explored text classification. Specifically, they addressed the type of primary textual records or context (such as identity documents). The method proposed by the authors locates the text and defines its class. The classification is done considering the nature of the document, country of citizenship, edition, and the noticeable side (main or back). They showed that this strategy effectively classifies and retrieves identification documents using just one sample as the source image instead of using a big training set. Coarse picture keypoints relate the document models to one image. Then, localizing and parsing are done to target and retrieve the text. The accuracy has been validated on real-world datasets. This job is quite tricky since it lacks consistent structural and textual details and numerous non-pertinent elements, e.g., picture, name, address, etc. Reference models are generated from primary source documents. The authors concluded that only one reference image is required to build a model. Better matching is done to identify the text and material extraction. About 3042 documents have been used to 64 groups and reached an accuracy of 96.6\%.
A match is made on all versions in the library. 

Vilas et al. \cite{vilas2018classification} created an image classifier to analyze similar topics and then applied their proposed framework to classify the ID documents. The authors arrived at the workflow after addressing the issue in many directions to help others in the machine learning field. They used several attribute extractor algorithms to find the most appropriate algorithms for document classification. They used virtual machines on the cloud to process the features at 16 images per second. Then, the authors used a neural network architecture and hyperparameters to optimize a convolutional neural network, and 98\% precision was obtained.

Ngoc et al. \cite{ngoc2018saliency} investigated the programmed separation of identity documents in smartphone pictures or videos using visual saliency (VS) to evaluate many VS styles and determine which one works well. To achieve this, the authors suggested a new VS system on a current distance fitting the possibility of scientific geomorphology. VS-based methods for computer vision had not been considered as of the time of this research. They presented the following visual saliency maps viable with advanced approaches, which showed potential identity and segmentation, even though contents were not considered. They also executed real-time on smartphones.  The authors attempted to expand the Dahu color space to make object detection more feasible. They considered an image as a four-sided entity with no boundaries. The most important finding of the authors is that visual significance affects detection. The authors could finish the task in less time and improved performance because they could process the data faster and more efficiently.

The identity records of essential security concerns were the focus of this study. Sicre et al. \cite{sicre2017identity} considered this issue a top priority to image classification, a problem that drew much attention from scientists. Their goal was to see significant gains, deep learning processes, strong conversion skills, and positive results. They performed the picture classification by utilizing the BOW pipeline. The SIFT keypoint extraction requires either finding dense or discriminating among interest points. Compact extraction has to give an improved presentation in arrangement, while awareness ideas are revolution invariant. The authors performed primary experimentations on a dataset of 9 modules of French ID documents (FRA). This number included 527 teaching and research photos that fell in the 26-to-136 range. An extended dataset (Extended-FRA or E-FRA), with a limit of 2399 photos (86 to 586 total documents), was used. The last dataset comprised 446 samples. The authors extracted BOW, VLAD, and SVM-class attribute vectors and used them to train SVMs. Furthermore, they calculated CNN attributes from pre-trained grids. Descriptors were calculated with numerous grids, sheets, alignments, and gauges. Lastly, a VLAD aggregation of initiation maps across alignments and gauges was planned, and related picture descriptors are used to classify SVMs. The authors found that CNN geographies derived from pre-trained grids are easy to compute and effective \cite{sicre2017identity}.

Almaksour et al. \cite{almaksour2016classification} proposed a system for classifying images of ID documents on a complex background. They evaluated the design on the problem of categorizing French identity documents. They considered the system on the issue of classifying French identity documents.
This pairing is made between the points of interest detected on both photos with a distance metric based on local descriptors calculated around the dots. This global and coarse matching step, which aims to associate the document with a winning class, was followed by a fine matching step, which allows the form to be precisely located in the image. The results showed excellent recall and accuracy rates.

Skoryukina et al. \cite{skoryukina2019fast} studied the simultaneous ID document and their projective distortion parameters. They considered two possible instances. Firstly, if you are using a mobile device, a video stream is registered for upload. The second case studies/considers scanned images on a server. For individual circumstances, the necessities were precise for the input information and processing speed. The widespread method was planned, which permitted explaining the problem in individual cases. The process was constructed on the representative image as a collection of feature keypoints and their descriptors. However, to achieve supplementary precise alteration restrictions approximation, some of the specific lines were removed from the input image and used as supplementary topographies. The authors defined the procedures to associate corresponding feature facts, bars, and quadrangles to symmetrical confirmation using RANSAC. To evaluate the quality, a publicly available dataset, MIDV-500 was used.

Castelblanco et al. \cite{castelblanco2020machine} have explained that mobile enrollment in services like banking is gaining in popularity. It is common that a photo ID is requested in these processes. It is crucial to verify the fundamental text recognition features for this to be successful. Furthermore, inspiring circumstances might be identified, for instance, several backgrounds, varied light eminence, viewpoints, standpoints, etc. In this study, Castelblanco et al. \cite{castelblanco2020machine} defined a machine learning method for images of records. They suggested a document authentication pipeline, which involved different research modules and visual features in evaluating document form and legitimacy. The authors used Colombian passports to test the methodology. The study showed that complete enrollment processes were successful. The machine learning system gave an accuracy of 98.4\%. The brand classification method had a 97.7\% hit rate and an F1 score of 0.974.

Tensmeyer et al. \cite{tensmeyer2017analysis} explain, the CNNs are advanced representations for ID document image classification responsibilities. Though, several methods depend on constraints and manners for categorizing natural images, varying from document images. They investigated the CNNs to see if it is proper and then performed an analysis to find out if this approach results in high accuracy in ID document classification. Among other consequences, the authors surpassed the advances on the RVLCDIP. They also examined the cultured topographies and discovered that CNN qualified on RVL-CDIP acquire region-specific design topographies. The highest advanced output on RVL-CDIP was observed at 90.8\% precision. Using CNN, the authors observed proof on RVL-CDIP that CNN is studied intermediate interface features. The neurons' exact positions are related to their components' layout (graphical, text, or handwriting).

Indonesian Electronic ID cards, which are the data that have been analyzed in \cite{satyawan2019citizen} have been generally accepted and used since 2011. Issues such as extracting the ID document text data and detecting the fields of ID cards are the hardships  Satyawan et al. \cite{satyawan2019citizen} have faced. The authors proposed their model to overcome the issues mentioned above. In their study,  they used image analysis and optical character recognition (OCR) to find electronic identification cards. The authors carried out the testing on the dashboard used by a car company. They associated gray-scale pre-processing methods with binary image processing systems such as Sobel, morphological transformation, and OTSU. Text area extraction practices a kernel that recognizes the NIK text range and name on the ID card. The experimentations with training data were completed using the tesseract 4.0 display. The authors observed that they could achieve almost 98\% precision by using their ID and optical character recognition techniques.

The growing usage of open-source libraries for image recognition processes utilizing machine learning algorithms contributes to datasets becoming more scalable. There is a degree of difficulty in rendering an image. This is due to the number of colors, objects, figures, or signs, which complement the considerations of image classification (point of view, lighting, deformation, occlusion, intraclass variation, and background clutter). When applying machine learning algorithms with deep-focus architectures, complexity is reduced. It is because these approaches allow increasing accuracy by creating a significant number of activation functions that mitigate these considerations, and with the use of multiple layers combine their responses.  Where images are used as inputs, image classification can also be seen as a solution that provides security for applications. 
Millam et al. \cite{millan2018comparative}  tested the precision of image classification methods using machine learning in an unusual situation, such as organization recognition, incorporating both text and picture data. The research highlighted and recommended the Transfer Learning/ Image Retraining technique for allowing the training and validation phase to be carried out efficiently and reduced times related to the other methods, offering greater certainty of the final classification by returning the accuracy obtained for each class existing in the classification. The dataset had a high architecture, varied community, and dense population. They observed that for this particular case study of image classification, it was necessary to implement a previous classifier debugging the images that enter the classifier resulting from this research.

Kumar et al. \cite{kumar2013unsupervised} displayed a learning-based method for calculating operational resemblances among document images for un-supervised examination in huge document groups to generate a better future category set. The process was constructed on numerous stages of satisfaction and structure. A bag-of-visual terms similarity calculation works well on SURF features at a local level. They recursively segregated the document, and a histogram of codewords is calculated for each panel. The operational resemblance is produced with a random forest classifier accomplished through these histogram topographies. The authors used three various databases for evaluating variously-sized documents and structurally similar documents. They found out that the planned method can confine the precise numeral of modules in the NIST tax-form and table datasets. For photos with only a design match, the layout and content were different, and grouping heuristics might be needed.

\subsection{Crowdsourcing as Post-Processing}
In some cases, the classification accuracy is significantly critical and considerable. Identity document classification is one of the image classification categories in which the classification results must be as accurate as possible, especially in some research such as the current study that the results of the classification phase feed the ID document authentication phase. Hence, the state-of-the-art in applying crowdsourcing to the results of the different image classifiers, as a post-processing approach for image classification accuracy enhancement, has been analyzed in this section.

De Herrera et al.~\cite{de2014crowdsourcing}, describe the ImageCLEFmed benchmark proposes a biomedical classification challenge that automatically determines imaging modality from biomedical journal papers. As a result, a small number of image forms are underrepresented, posing a problem for automated image classification. The training set manually checked using Crowd-flower. This website enables the usage of additional individuals and the payment of crowdsourcing fees, and the unrestricted use of personal contacts. Crowdsourcing necessitates stringent quality management or the use of trustworthy individuals~\cite{DBLP:conf/caise/BeheshtiVBT18,DBLP:conf/colcom/AllahbakhshIBBBF12,DBLP:conf/momm/GhafariJBPYO19}.
Nonetheless, it will rapidly have links to many courts, thus improving multiple machine learning activities. The results indicate that manually annotating many biomedical images will aid in image classification~\cite{de2014crowdsourcing,DBLP:conf/momm/RezvaniBT20}. This work aimed to use crowdsourcing to enhance the efficiency of automated classification.

Zhao et al. \cite{zhao2017range} explain that their proposed crowdsourcing approach enables humans to solve issues that machines are incapable of solving, such as emotion analysis and image recognition. Crowdsourcing has the potential to improve the accuracy of medical picture recognition. They took a hybrid approach to image recognition using a computational algorithm and a crowdsourcing method. Furthermore, it is irrelevant to enhance the precision of these classification algorithms. They suggested a hybrid for resolving the problem, which achieves a higher accuracy level than just classification algorithms. Simultaneously, it processes only photos for which classification algorithms fail, resulting in a lower monetary expense. They developed an efficient algorithm for generating a range threshold for crowdsourcing their classification algorithm inside the system. Experiments demonstrate that their approach will increase medical picture recognition quality while still lowering financial expenses.

He et al. \cite{he2019crowdsourcing} discuss two different worker methods: the Worker Quality Evaluation Model (WQEM) and the Worker Performance Prediction Model (WPPM) to ensure crowdsourced responses and accuracy. Due to the absence of a crowdsourcing framework for health data retrieval, health image arrangement findings are challenging to obtain. Thus, this article introduced a crowdsourcing stage for remedial image arrangement. They examined the description of medical images in this article. This article optimizes the classification process from job selection to worker selection by using crowdsourcing to solve medical image classification helping the HMC system determine which photos need re-marking by crowdsourcing staff, significantly reducing the load and stage costs.

Hsing et al. \cite{hsing2018economical} demonstrated while camera trapping to track mammalian wildlife generates massive picture data sets that must be categorized. As a result, a movement toward crowdsourcing picture recognition has emerged. Numerous categories can be obtained for high-profile studies of charismatic faunas, allowing for consensus evaluations. However, demand for crowdsourced classifications may exceed supply in communities. MammalWeb is a local cable network in North East England that engages resident scientists in the apprehension and variety of camera trap picture sequences. The authors demonstrated that the likelihood of correctly classifying an image sequence approaches 99\% for approximately concordant crowdsourced categories within each in their global image sequences pool. However, there was considerable overlap between species.

It was shown by Li et al. \cite{li2020rsi} that in remote sensing image processing, the classification of images plays a critical role. Although the area continuously ignores an extensive benchmark comparable to ImageNet, In current centuries, deep convolutional neural networks (DCNNs) have achieved substantial raw image recognition revolutions. They proposed a remote sensing image classification (RSI-CB) using crowdsourced data. The claimed land features in sensing images can be explained with the facts of attention by using crowdsourced data. The authors developed an extensive global standard for sensing picture arrangement utilizing this approach. This standard is widely distributed geographically and has a high total picture count. Divided into six divisions of 35 sub-categories and features over 24,000 photographs with 256 by 256 pixels. They performed an arrangement trial on many standard deep learning grids, demonstrating that their arrangement precision was superior to that of different data sets due to their increased altitudinal determination.

A large number of gray-scale photos of the Smoky Mountains have been gathered in the University of Tennessee Library. The presented system of defining and marking photographic features by Simpson et al. \cite{simpson2016tagamajig} is manual. For example, "Floating Forests" with users tasked with identifying and labeling kelp forests in satellite photographs. At the same time, Ancient Lives enlists their assistance in transliterating 2000-year-old texts that Oxford University scholars have been unable to read for over a century. Addressing this, they recommended making picture collections accessible to the community over a network application. The application would appeal to outside tourists, preservationists, and experts acquainted with the area, like geologists, rangers, and historians interested in naming the more identifiable photographs. Users will mark landmarks with the help of a hierarchically organized data collection of landmark names, accessible through an incremental quest. The authors mentioned that with adequate involvement, the picture set might be easily classified and labeled beyond what is actually possible with the library's small staff. 

Saralioglu et al. \cite{saralioglu2019use} explained that the term crowdsourcing describes the process of using individuals to address a particular issue, typically concentrating on research techniques that minimize the time, expense, and effort required to generate results. By definition, crowdsourcing asserts that communities can make choices more intelligently and accurately than the most knowledgeable person within them. They examined the feasibility of using crowdsourcing to gather control points useable in an assessment that was conducted with the class values of 1000 arbitrarily created control points. This experiment aimed to determine the precision of unique class values arrived by three separate users using the majority elective system.


\section{Methodology}

In this section, the required methods used in this project's construction and implementation process of this project are explained. Also, basic concepts and methods that are directly or indirectly related to the research topic are presented. This literature and explanations have been used in the implementation of the studied model. In addition, the variables, and the dataset which has been used and analyzed in each model, have been explained. Additionally, the implementation and the results of the failed approaches have been touched on.

\subsection{Dataset Characteristics}

The dataset intended to be classified in the current study is a specific type of imagery dataset with some unique characteristics, which varies the classification approach with usual image classification approaches. The mentioned characteristics are as follows:

\begin{itemize}
\item The number of the target classes is high due to the variation of the ID documents worldwide.
\item Due to security and privacy considerations, ID Document images are rarely accessible, so the method should be capable of getting trained by just one training data.
\item Many similar ID documents should be classified. The unavoidable similarity of the ID documents is one of the main challenges of the ID document classification. For instance, as mentioned in this section, several driver's licenses in each state of Australia are almost similar, and their difference is mostly in few text features and/or their color. Some of the NSW (New South Wales) driver's licenses are shown below. (Figure \ref{fig:03-15} illustrates some of NSW driver's licenses)

    \begin{figure}[h]
    \centering
    \includegraphics[width=0.8\textwidth]{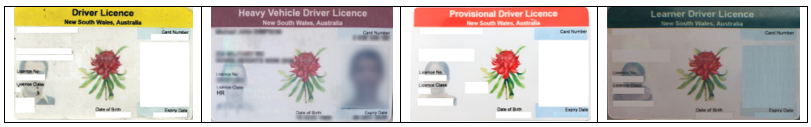}
    \caption{NSW driver's license samples.}
    \label{fig:03-15}
    \end{figure}

\item The results of the identity document classification will feed another method to authenticate the ID documents based on the pre-identified type of the document with the classification phase. Therefore the results of this phase of the project indirectly affect the authentication phase. Since document authentication is intrinsically crucial, the ID document classification must be done as accurately as possible.
\end{itemize}
Due to the mentioned characteristics of the dataset, the methods that are more efficient on smaller datasets should be considered. Thus, the feature detection and matching methods have been used to compare a sample image to the source images to calculate the similarity of the images. Finally, the most similar two images have been considered the same class or type of ID document.

\subsection{Approach 1 (HOG + Color-Name + SP3)}
The characteristics of an identity document imagery dataset led us to search for the approaches that have worked on the same kinds of datasets, dealing with a high number of classes, and working on a smaller dataset, comparing to the large datasets, which enable us to train machine learning models. Therefore, as the first step, a proposed method mentioned in the background section, in which the authors have reported 97.7\% accuracy on a global ID document dataset of 75 classes, has been replicated. The method works based on a feature similarity check between a source images and the test image. Source image-set contains one high-quality image per class, and the sample or test images have been compared to all source images with the proposed method. And finally, the class of the source image with the highest similarity has been accepted as the target class \cite{simon2015fine}.

\subsubsection{Methods Description}
As discussed above, the mentioned approach suits the current use case the most due to classifying many classes based on just one sample per class with high accuracy. The detail of the approach is explained in this section. In addition, for more understanding, a brief explanation of each method has been indicated before each step explanation.

\textbf{HOG (Histogram of Oriented Gradients)}
HOG is an often-used method for extracting features of the images. It is commonly used for target recognition in computer vision. In a 1986 patent filing, Robert K. McConnell of Wayland Research Inc. represented principles of HOG without using the word HOG. The histogram of the directed gradients descriptor is based on the idea that the intensity gradients distribution or edge directions can define the appearance and form of local objects within a picture. Images are subdivided into tiny-linked regions called cells. A HOG is constructed for each pixel within each cell. Concatenation of these histograms is used as the descriptor \cite{dalal2005histograms}. HOG is capable of supplying both the orientation of the edge and its characteristics. This is accomplished by extracting the edge's gradient and orientation.
Additionally, these orientations are determined in 'localized' segments. This is accomplished by segmenting the whole picture into smaller regions and calculating the gradients and orientation for each region. Finally, the HOG will produce an individual histogram for each of these areas. Histograms are constructed from the gradients and orientations of pixel values \cite{naiemi2019efficient}.

A HOG feature detection has been applied to the current study's dataset and created a vector containing the HOG features in each cell of the vector. To classify the sample HOG feature vectors, the cosine similarity approach that operates by calculating two vectors' similarity has been used. Cosine similarity determines the similarity of two vectors by detecting the similarity of the directions of the two vectors. Testing HOG-based classification on the initial dataset, including 105 data, ended up with 44 misclassification and 58.09\% accuracy.

\begin{figure}[h]
    \centering
    \includegraphics[width=0.5\textwidth]{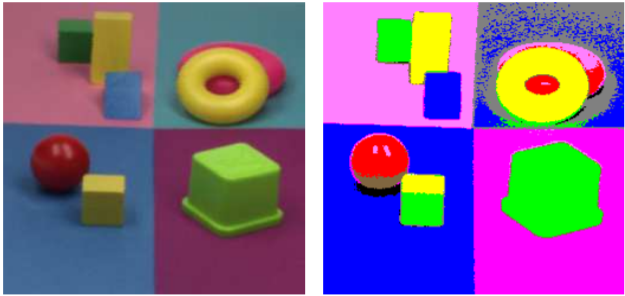}
    \caption{Color mapping to limited colors \cite{vanapplying}.}
    \label{fig:03-16}
    \end{figure}

\textbf{Color-Name}
The next step proposed by \cite{simon2015fine} is to calculate the color-name vector of the images. The color-name means to map all colors available in an image to a small number of detectable colors with naked eyes and usually are the primary colors. As the mentioned paper authors suggested, in the current study, the model proposed in \cite{vanapplying} has been followed. The eBay auction website\footnote{https://www.ebay.com.au/help/selling} images have been used. All RGB(Red, Green, Blue) color-space colors have been mapped to 11 primary colors (black, white, blue, yellow, brown, gray, orange, green, pink, red, and purple). The used dataset contained 440 images, 40 images per color. The mean value of each 40 sample color of a specific color has been accepted as the unique value which represents the relevant color. Figure \ref{fig:03-16} shows how the different color values of a particular color have been mapped to just one value.\\

Then the distribution of each color available in each source and sample image has been calculated. After that, an 11 cells color vector has been created, in which each cell of the vector shows the distribution of the corresponding color. Then, the color vector of each source and sample image has been added to the relevant HOG vector and calculated the cosine similarity between each sample image and all source images. Like the previous method, the highest similarity has been accepted as the target class. This time the model resulted in 40 misclassified images and 61.9\% accuracy.

\begin{figure}[h]
    \centering
    \includegraphics[width=0.5\textwidth]{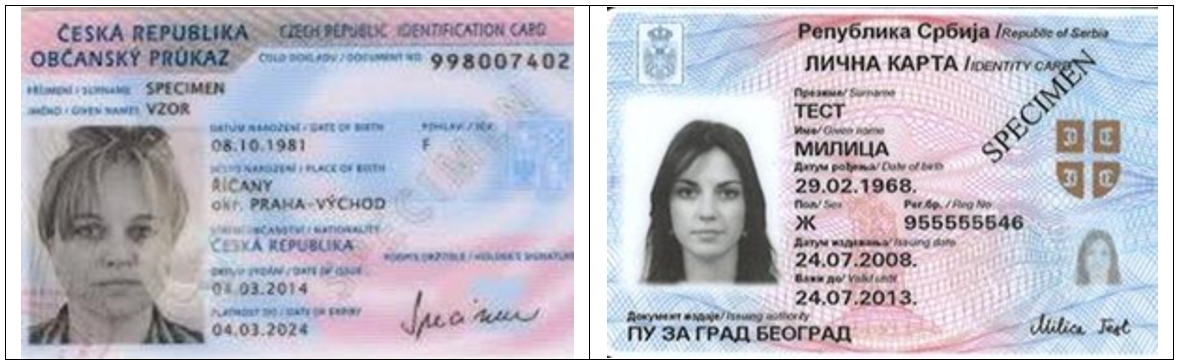}
    \caption{Same color distribution with different color pattern.}
    \label{fig:03-17}
    \end{figure}

In some documents the color distributions are almost similar, but the color diffusions or color patterns are different. Figure \ref{fig:03-17} shows one sample of the mentioned ID documents. However, the following samples do not exist in our dataset.

\begin{figure}[h]
    \centering
    \includegraphics[width=0.7\textwidth]{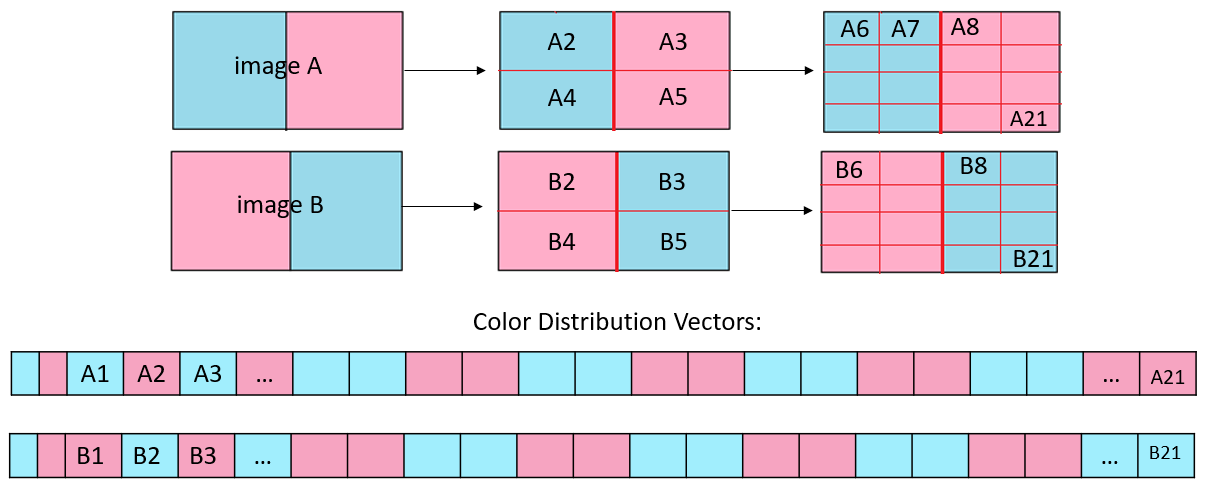}
    \caption{Same color distribution with different color pattern.}
    \label{fig:03-18}
    \end{figure}

\textbf{SP3 (Spatial Pyramid 3)}
The method mentioned above cannot distinguish between such images, so as proposed in the fundamental paper \cite{simon2015fine}, the SP3 (Spatial Pyramid 3) technique has been applied to eliminate model confusion facing these kinds of data and possibly enhance the accuracy of the fusion model. Applying the SP3 technique, the color distribution vector has been calculated 21 times for different pieces of the image. First, the color distribution of the original image has been calculated, then divide the original image into four sections and calculated the color distribution for each of those four sections. Finally, each of the previous pieces of images has been divided into four new sections, and the distribution of each 16 new sub-sections has been calculated. Then a 1+4+16 =21 cells vector has been created, in which each cell includes 11 different cells indicating the distribution of each of our intended colors in the corresponding section. Figure \ref{fig:03-18} illustrates how SP3 works.

\subsubsection{Conclusion}
Applying the SP3 technique to our source and sample images, we got a 2 percent enhancement in the fusion model accuracy. Hence the final accuracy of the mentioned method was 63.8\% which was far from our expectation. To enhance the model's accuracy, all the train and test data were analyzed manually to determine why the reported accuracy in the fundamental paper could not be replicated. The outcome has been explained in the next session, in addition to the appropriate solution to overcome the issues.

\subsection{Approach 2 (HOG + Fuzzy color name)}
\begin{figure}[h]
    \centering
    \includegraphics[width=0.9\textwidth]{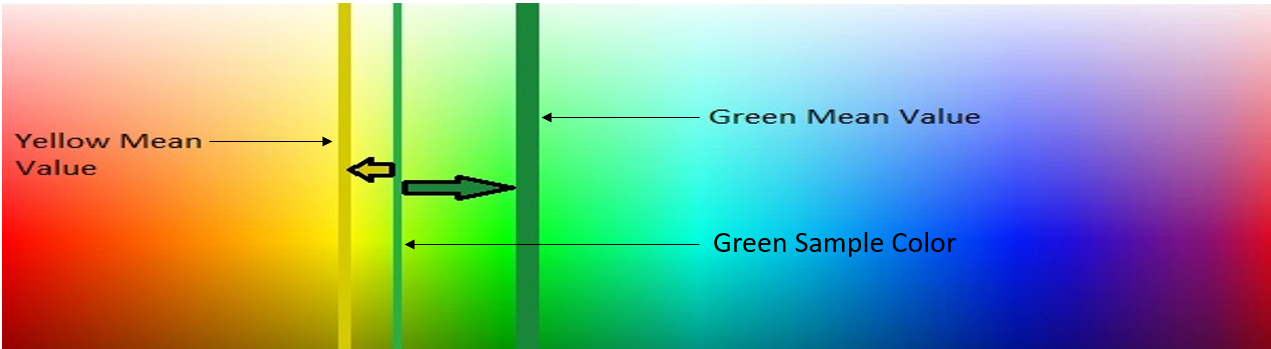}
    \caption{Mean value-based, color detection failure.}
    \label{fig:03-19}
    \end{figure}

Analyzing the previously explained dataset and the color detection method, it has been realized that the color detection has not been working well, i.e., the yellow color has been detected as an orange one. The explained approach is willing to map the mean value of a range of colors to all colors in the same category. The first reason of the approach failure is that if all color values are considered as a population, minimal numbers of instances of 40 colors cannot cover and represent the whole population. The second reason is that, even if the sample size of 40 covers all color regions, then the mean value will be the exact mean value of that region. The problem is that if a sample color comes in, it should be closer to the mean value of the target region to be detected correctly. In Figure \ref{fig:03-19}, the mean value of yellow and green color regions have been indicated by a line with the same color, assuming that the mean value derived from the 40 corresponding samples can represent the exact mean value of the color region. In the sample below, the middle green line represents the value of a sample color which we expect to be detected as green, but because the sample color distance is closer to the mean value of yellow, compared to the distance of the sample color to the mean value of the green color, the sample color will be detected as yellow instead of green. The explained example shows the reason why the mentioned color detection method fails in many color regions.

\subsubsection{Methods Description}
\begin{figure}[h]
    \centering
    \includegraphics[width=0.2\textwidth]{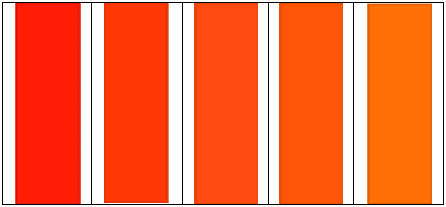}
    \caption{Uncertainty in Color Recognition.}
    \label{fig:03-20}
    \end{figure}

To overcome the above issue, a Mamdani fuzzy color detection has been proposed, not only to be able to cover the whole color space but also to detect the color as accurately as possible. A Mamdani fuzzy system is proposed because uncertainty in color values has been detected, especially when it comes to the borders of the colors. For example, naming the below three colors, shown in Figure \ref{fig:03-20} some people may believe all of them are red, while some others may consider 1 or 2 of the below colors like orange.

\begin{figure}[h!]
    \centering
    \includegraphics[width=0.85\textwidth]{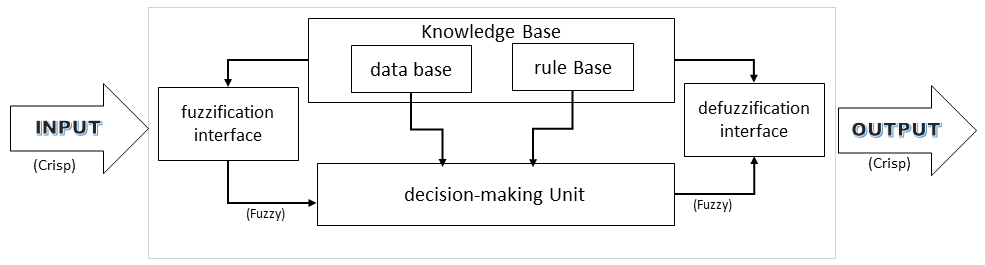}
    \caption{Fuzzy System Structure.}
    \label{fig:03-Fuzzy}
\end{figure}

\textbf{Fuzzy Systems}
Professor Lotfizadeh introduced the fuzzy theory in 1965 in an article called Fuzzy Sets. The first step for building a fuzzy system is to obtain several if-then-fuzzy rules. If the behavior of the expert is apparent, the resulting knowledge is conscious (structured, directional, and wise); otherwise, it will be unconscious (or intuitive). Conscious knowledge can easily be turned into a set of fuzzy if-then rules, in which case the system will not need further expert intervention. However, expert knowledge is still needed to measure and validate the input/output pairs if the knowledge is unconscious. Finally, in both cases, the system input, which is a set of if-then rules, is injected into the fuzzy system. System inputs are generally in the form of definite inputs, so the system needs to fuzzy the inputs in the next step. Fuzzification is performed for those inputs that do not follow fuzzy rules. Figure \ref{fig:03-Fuzzy} demonstrates the general structure of the fuzzy system. As shown in Figure \ref{fig:03-Fuzzy}, the definite inputs are fuzzy using the system knowledge base, and after processing by the fuzzy inference unit, they are sent to the decapsulation module to produce the definite output.

\begin{figure}[h]
    \centering
    \includegraphics[width=0.8\textwidth]{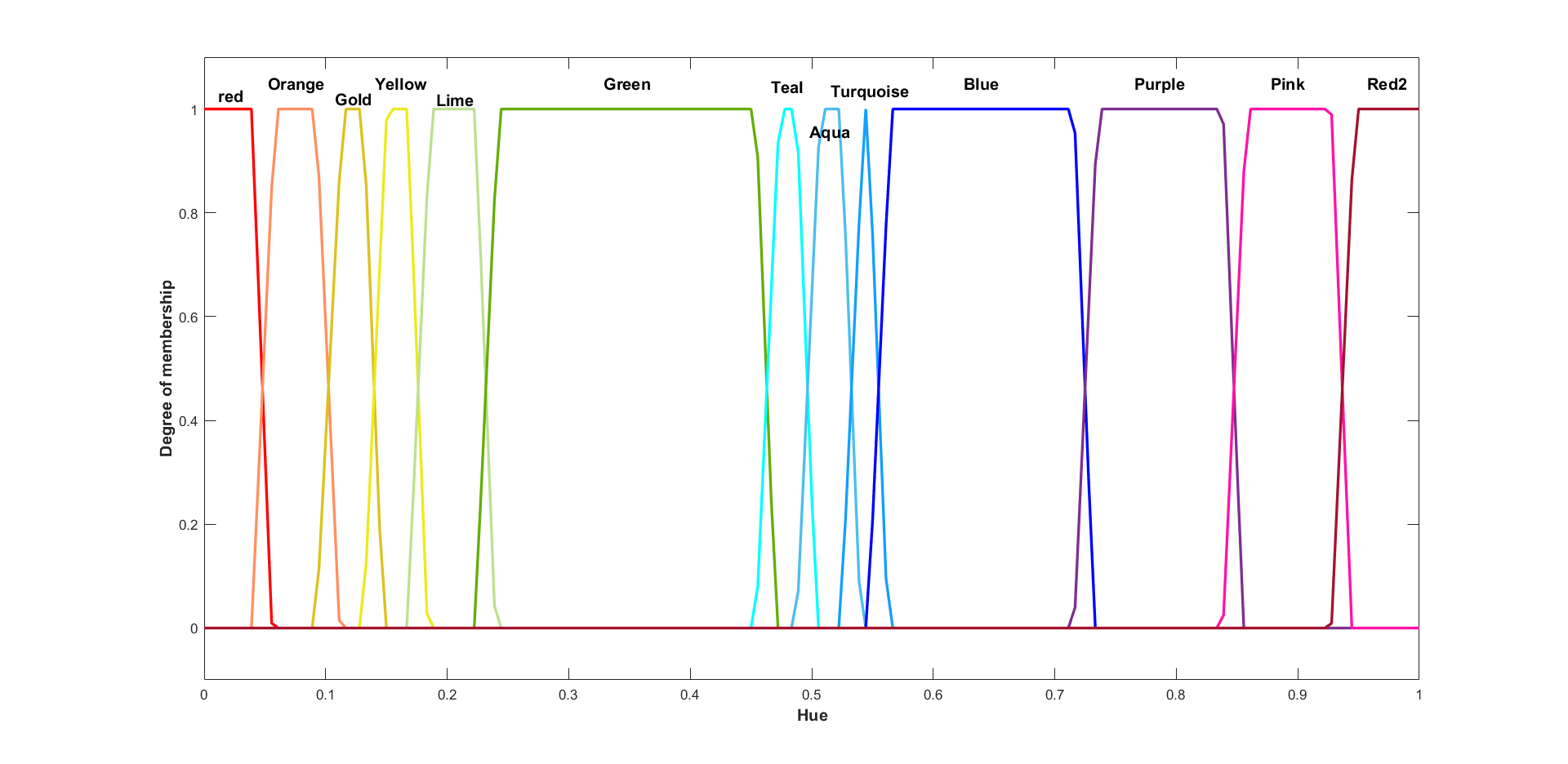}
    \caption{Input 1 - membership functions.}
    \label{fig:03-Hue}
\end{figure}

\begin{figure}[h]
    \centering
    \includegraphics[width=0.8\textwidth]{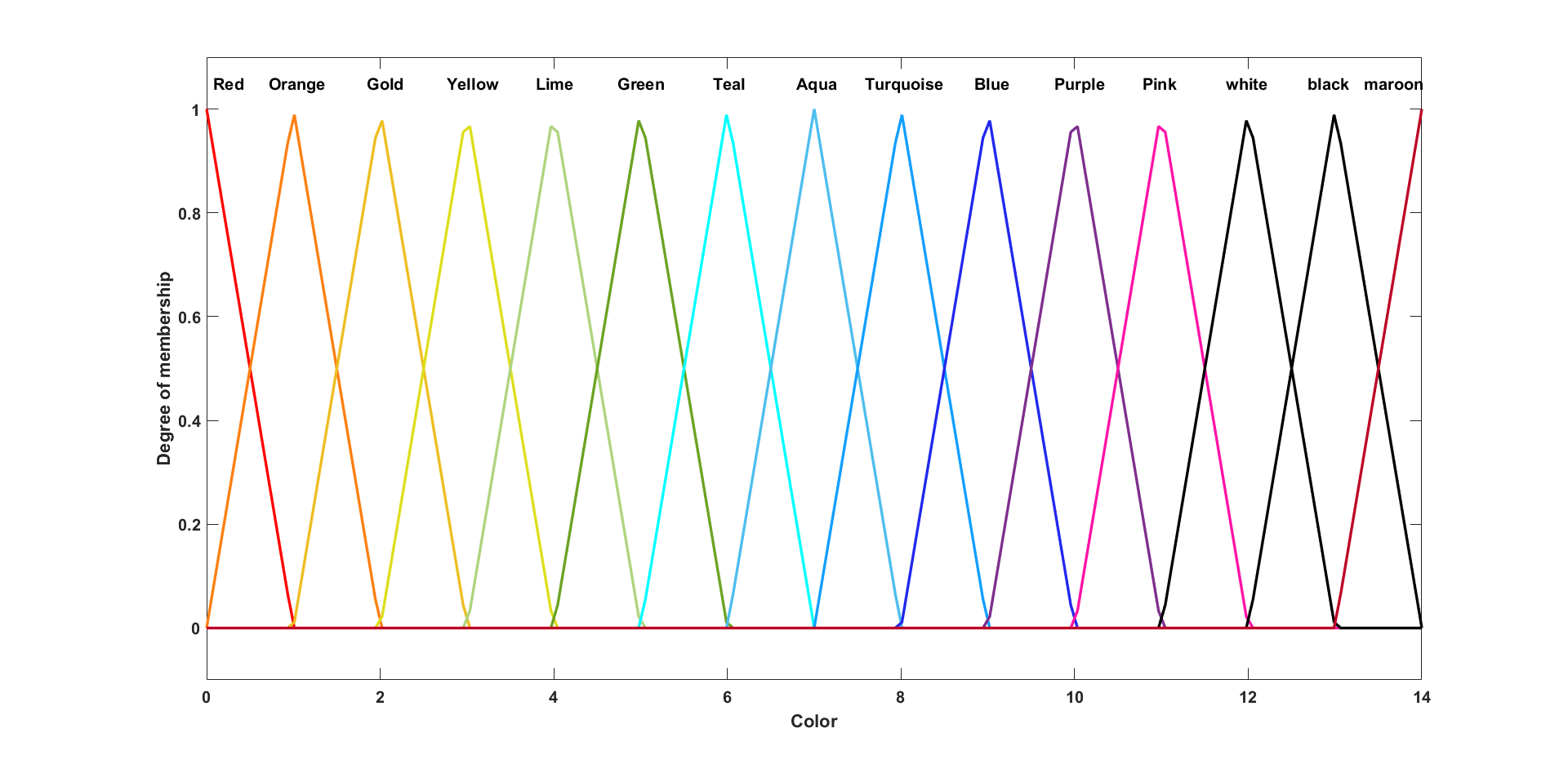}
    \caption{Output - membership functions.}
    \label{fig:03-output}
\end{figure}

According to the current imagery dataset, the whole color space has been mapped into 15 colors. The available ID documents. To consider and minimize the effect of the light intensity and shadow on the images, the HSV (Hue, Saturation, Value) color space has been used. Therefore the proposed fuzzy system includes three inputs of hue, saturation, and value of the color. Hue varies between 0 to 365, whereas saturation and value are limited in 0 to 100. To integrate the three input variables, the input variables have been normalized between 0 and 1 using the min-max normalization method. The trapmf(x, params) function has been used to compute the input fuzzy membership values. Also, the trimf(x, params) has been used to calculate the output membership functions. Figure \ref{fig:03-Hue}, shows the membership functions of input 1, which belongs to Hue, and Figure \ref{fig:03-output} represents the membership function of the system output.

\textbf{Fuzzy Rules}\\
In the current system, due to the fuzzy nature of the output variable, the antecedent and consequent of all generated rules must also contain fuzzy linguistic variables. The fuzzy rules of the system have been extracted based on the colors of the ID documents existing in our dataset and considering how the combination of HSV values can create different colors. The weight of all input variables and fuzzy rules are considered the same, but the rules are written in such a way that some variables are more effective than other variables when they contain certain values. Finally, according to the color's characteristic and the highest probability of a color occurrence in the middle of each linguistic interval, centroid-de-fuzzifier was chosen as the most suitable de-fuzzifier. The model ended up with 54 fuzzy rules that resulted in the expected color detection.

\begin{figure}[h]
    \centering
    \includegraphics[width=0.6\textwidth]{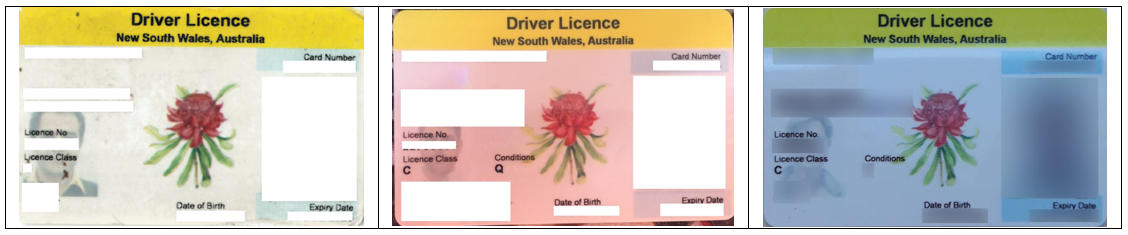}
    \caption{Lighting condition effect, on the color of the same type of ID documents.}
    \label{fig:03-21}
\end{figure}

To evaluate the fuzzy color detection model, 20 random color values have been selected from each of the predefined color-names. They have been tested through the model to see if the model can correctly return the color name. According to the results, the model has been detecting all the 300 selected colors correctly. Therefore, the color-name vector explained in the previous section has been created with the results of the fuzzy color detection method to see if the overall result of the classification improves. Applying fuzzy color detection to the mentioned fusion model resulted in achieving a 10\% accuracy improvement. However, the overall accuracy is still far from the current study's expectations. Analyzing the dataset, the samples shown in Figure \ref{fig:03-21} have been faced. Although the samples belong to the same class, affecting different lighting conditions, the ID document color has been captured in three different colors.

\subsubsection{Conclusion}
Analyzing the results of the previous section, it has been realized that the color detection method has been working as accurately as expected. Therefore, in this section, we have tried to enhance the accuracy of the overall mentioned fusion classification model by improving the results of the color detection method using a fuzzy system color detection. Although the proposed color detection model has been working accurately, a significant enhancement in the classification model has not been returned. Analyzing the dataset, it has been figured out that many samples in the same class are different in color, depending on the image capturing conditions. So it can be concluded that using the color attribute is not a good way to classify the ID documents.

\subsection{Approach 3 (SIFT + Feature Engineering)}
\begin{figure}[h]
    \centering
    \includegraphics[width=0.5\textwidth]{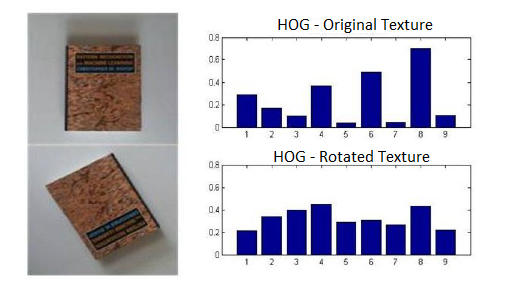}
    \caption{HOG results on original texture vs rotated texture \cite{cheon2011rotation}.}
    \label{fig:03-22}
\end{figure}

Taking the method proposed in the previous section, it has been realized that the accuracy of the fusion model is not as expected. Furthermore, it has been figured out that due to the lighting conditions, different colors after capturing images are detected for the same images. The mentioned reasons concluded that color-based classification was not a good idea for the ID document classification. In addition, as discussed previously, the HOG feature detection is very rotation sensitive. If the image of the same object gets captured with different angles, a different histogram of oriented gradients is returned. Figure \ref{fig:03-22} shows two different HOG results of the same object captured in two different rotations.

\subsubsection{Methods Description}
Different histograms are calculated because this method works based on the orientation or angle of the gradients, and image rotation directly affects the results. The current study dataset includes different data with various rotations and lighting conditions, which also is possible in the real world; furthermore, manually adjusting the rotation and applying pre-processing on each customer document is not possible and rational. Hence, it has been decided to use another feature detection model, which is rotation, scale, and orientation invariant.

\textbf{SIFT (Scale Invariant Feature Transform)}
SIFT, is a computer visualization feature recognition process for detecting and reciting local structures in pictures. First, the SIFT key themes of items are pulled out from a collection of reference images and deposited in a database. The recognition of an entity in an innovative duplicate is accomplished by comparison each feature in the novel duplicate to the database and recognizing identical applicant structures based on the Euclidian gap between their feature routes \cite{lowe1999object, lowe2004distinctive}.

Every item contained inside an image may have exciting points to deliver a "function definition" of the entity. This definition, taken from an exercise image, will then be used to classify the entity in a test image that contains several other items. To perform accurate identification, the features derived from the training images must remain detectable as the image size, noise level, or lighting varies. These points are often located in areas of the image with strong contrast, such as object edges \cite{lowe1999object, lowe2004distinctive}. Additionally, these elements should maintain their relative locations in the initial scene from image to image. Correspondingly, structures embedded in expressed or supple items are stereotypically inoperable if their interior geometry changes between two pictures in the processing set. On the other hand, the SIFT identifies and utilizes a significantly greater range of features from an image in operation \cite{jin2017tracking}. Being a robust object identification leads the SIFT to be the most popular algorithm for extracting an image's critical and unique features. The SIFT model is invariant to rotation, scale, brightness, positioning, and radiance shifts, also partly invariant to affinal alteration and other imaging parameters \cite{kasiselvanathan2020palm}.

According to the mentioned characteristics of the SIFT model, this feature extraction model can be an extremely proper approach to extract the unique features of our source and sample images. Therefore, in this section, a source/train dataset has been created in which one of the highest resolution samples of each class among all the available samples has been collected. Then applying the SIFT model, the SIFT keypoints and descriptors of each source and sample images have been extracted. The Brute-Force matcher has been used to match the features, which simply compares every descriptor of a specific feature of the sample image with all other descriptors in the source image using a distance calculation method and returns the closest one as the matched feature. Finally, the class of the source image has been defined based on the number of matched features. Applying this method to 152 sample images, 96.05\% accuracy has been returned.

In spite of the SIFT good results, to enhance the accuracy, the dataset has been analyzed again. It has been tried to find out the unique features of the dataset and eliminate the non-informative features. In fact, the feature engineering concept has been used to enhance the accuracy of the classification model.

\textbf{Feature Engineering}
The feature has been around for many years. Despite trying to define features more precisely, the term is misused in several ways. A feature is the functionality of a cohesive system. Every candidate sees this set differently. It is not entirely clear which definition is the best, however. Because features are created in the problem and not in the solution, the first appears to be more beneficial. Additionally, explicit artifact groupings can be found in the second definition, based on lifecycle artifact models that employ the first definition.

The aim is to build a robust conceptualization of a feature and, more importantly, future implementation of the features. This is the feature engineering area. The primary goal of the feature engineering goal is to ensure that features are at "usable" quality levels in all the whole life cycle activities. Additionally, features include elements that must be identified in specifications, evaluating design implementations based on their ability to incorporate these and feature interaction, and understanding feature con features to structure meaningful relationships over the entire lifecycle.

Feature engineering holds the potential to leverage and increase other software-requirement efforts over the program's lifetime in a highly structured manner \cite{turner1999conceptual}. Hundreds of distinct variables or features have been used in machine learning and pattern analysis in the past. Several tools are used to counter burdensome and wasteful variables. Feature Selection reduces the curse of dimension and improves the accuracy of prediction. To find several specific features to reduce the model's generalization error. The goal of feature selection is to filter out noisy or irrelevant input while retaining meaningful results \cite{guyon2003introduction}. An irrelevant feature removal must be carried out for each class or feature to measure its relevancy. From a machine learning perspective, irrelevant variables lead to the problem of poor generalization \cite{chandrashekar2014survey}.

\begin{figure}[h]
    \centering
    \includegraphics[width=0.8\textwidth]{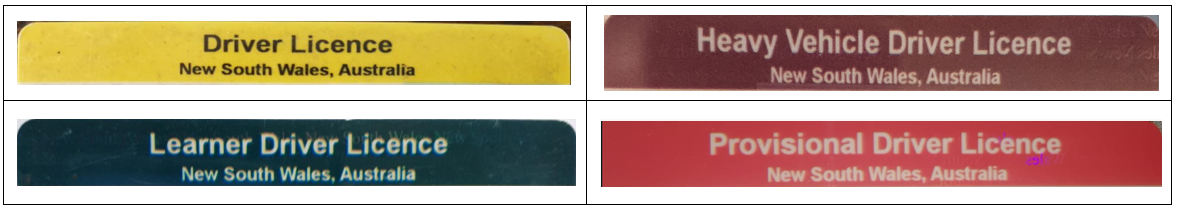}
    \caption{Top banner samples of NSW driver's licenses.}
    \label{fig:03-23}
\end{figure}

It has been identified that at least one visual or text feature exists in all of the documents that specifically identifies the type of an ID document. For example, as shown in Figure \ref{fig:03-23} the top banners of Australian driver's licenses include a written text that shows each document's type, and usually, this part of the document is used to feed the OCR models. In this section, the mentioned text features have been considered as visual feature not text.

This time, instead of selecting the whole document image, only a cropped part of the document which contains the unique visual or text features has been considered as training data. The SIFT keypoints and descriptors of each unique piece of the document have been extracted and matched the extracted keypoints with the whole sample image. Then the class of the piece of the document with the highest number of matching features with the sample image has been considered the target class. 88.8 percent accuracy has been reached by applying the explained method.

So far, two classification results for each class have been achieved. So a method is needed to combine the results of each classification to get a single result for each sample. In other words, the aim is to create a fusion model based on the results of the mentioned two classifiers. The result of each classifier is a number that indicated the amount of the highest matching features. The range of the matching features varies in different samples based on the image quality, resolution, and edges found in the image. Then it is required to normalize those numbers, to compare them with each other. Usual normalization methods, such as min-max normalization, is not helpful or functional in this situation. Because using the usual normalization methods, the distribution of the numbers cannot be identified. Whereas identification of the distribution model of the data after normalization was essential. Therefore it has been decided to use a standard score to normalize each classification matching result.

\begin{figure}[h]
    \centering
    \includegraphics[width=0.55\textwidth]{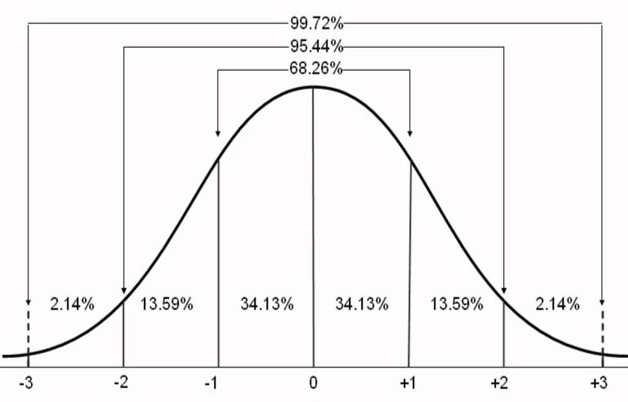}
    \caption{Z-Score in normal distribution.}
    \label{fig:03-24}
\end{figure}

\textbf{Z-Score}
The standard score or usually called z-score is the amount of standard deviations for which the significance of a raw score is greater than or less than the mean value of what is being evaluated. Raw scores above and below the mean value are considered as positive and negative standard scores, respectively. Hence, by calculating the Z-score, it can be analyzed how far any specific data is from the mean value of the distribution.
As demonstrated in Equation \ref{equ:03-01}, to calculate Z-score, the distribution's mean has been subtracted from each raw score of the population, and then the result has been divided by the population standard deviation. Standardization or normalization refers to the method of translating a raw score to a standard score. As shown in Figure \ref{fig:03-24}, Z-score range in normal distribution varies between -3 to +3 standard derivations. By calculating the z-score, the behavior of raw data in a population can be analyzed, a z-score shows where the data stands compared to the mean value of the distribution \cite{mendenhall2016statistics} (Equation \ref{equ:03-01}).

\begin{equation}\label{equ:03-01}
    Z = (x - \mu)/\sigma
\end{equation}

\begin{table}[h!]
\begin{tabular}{llll}
where: & & \\
Z = & standard score                   \\
x = & observed value                   \\
$\mu$ = & mean of the sample               \\
$\sigma$ = & standard deviation of the sample
\end{tabular}
\end{table}

The Z-score of each matching feature has been calculated to normalize them. Since one matching feature is assigned to each class, the number of matching features is equal to the number of the defined classes. As the next step, calculating the confidence levels based on the normalized data is needed to apply a proper logic such as fuzzy mean value to create a fusion model. One of the most popular methods to calculate the probability or confidence level of existence of the data in a specific class is the regression model. The output is a binary classification. In binary classification, it can be determined whether a sample belongs to a class or not. Therefore, a logistic regression, which is suitable for binary classification, can get the confidence score based on the normalized matching feature values.

\textbf{LR (Logistic Regression)}
In mathematics, the logistic (or logit) prototypical is used to estimate the likelihood of a specific type of case, such as pass/fail and on/off. This technique may be applied to prototypical in different types of activities, such as deciding if a picture includes a cat, dog, or lion. Each observed entity in the image will be given a likelihood ranging from 0 to 1, with a summation of 1 \cite{tolles2016logistic}. In LR's most straightforward procedure, it models a binary dependent variable. To estimate logistic model parameters, often logistic regression is used in a regression study \cite{tolles2016logistic}. A binary logistic methodically consists of a variable with two possible opposite values, i.e., lose/win, indicated by "0" and "1". In the logistic model, "1" will be assigned to positive classes or values, a linear combination of predictors, which are one or a few numbers of independent variables (binary or continuous), identifies the log of odds. A logistic function transforms log-odds to the probability that can vary between "0" and "1", which the value 1 represents the 100 percent probability of label "1" or class positive occurrence \cite{saberioon2018comparative}

By fitting a simple logistic regression on the calculated standard scores, a confidence score instead of each value can be generated, which shows the confidence level in classifying the sample. In other words, the confidence score shows the probability that each sample of data belongs to each class. Finally, applying a min fuzzy logic on the two confidence levels of each class (one confidence level for each classification model), an overall confidence level can be generated, and the class to which the highest confidence level belongs can be accepted as the target class. 98.02\% accuracy can be achieved by the deployment of the explained fusion model.

\subsubsection{Conclusion}
By analyzing the mentioned fusion model results, it has been decided not to use classification based on the image color. Also, according to the HOG feature detection sensitivity to rotation, it has been elected to use the SIFT model that implies a more robust feature detection model. A good outcome has been achieved by applying the SIFT model on the whole data. However, to improve the model's accuracy, it has been decided to apply the model to the unique text or visual features of each document.

Since the SIFT model does not return any confidence score and all returned is a value showing the number of matching features of sample and source image. A novel method has been proposed to combine the classifiers' results, mentioned above; the suggested model is based on statistics and machine learning. In the proposed method, the Z-score of each matching feature will be calculated, and then, to get the probability of each sample belonging to a specific class, logistic regression will be applied. Then applying the fuzzy min logic, the final confidence level will be calculated.

\subsection{Approach 4 (SIFT + OCR)}
\begin{figure}[h]
    \centering
    \includegraphics[width=0.45\textwidth]{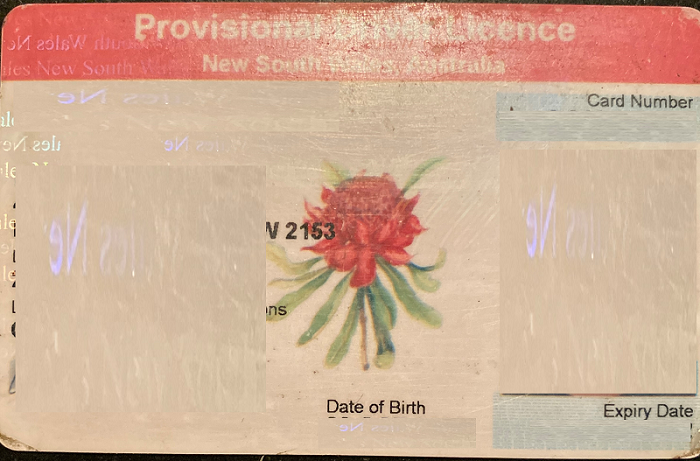}
    \caption{Spotlight effect on driver's license unique feature.}
    \label{fig:03-26}
\end{figure}

Although applying the proposed fusion model in the previous section enhanced the accuracy significantly, the following issues have been encountered which limits the use of the proposed fusion model:

\begin{itemize}
\item the target accuracy of above 99\% has not been achieved.
\item There are some ID documents, regardless of their colors, a unique visual or text feature that could identify the type of the document could be hardly identified. For instance, as illustrated in Figure \ref{fig:03-25}, the only feature which differentiates Queensland full driver's license from Queensland heavy vehicle driver's license is the word "Heavy vehicle", and the rest of the features are similar. Therefore, applying the SIFT model to find the rest of the features, the model usually returns more matching features with the full driver's license.

\begin{figure}[h]
    \centering
    \includegraphics[width=0.55\textwidth]{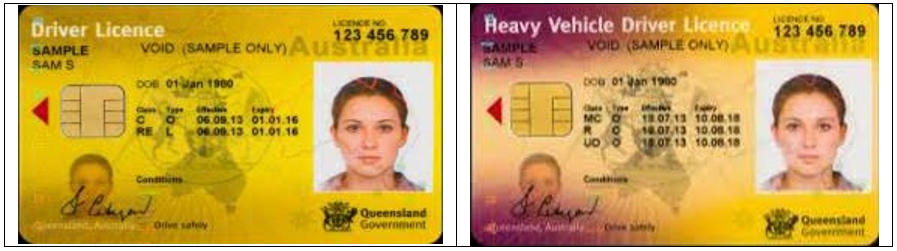}
    \caption{Very similar driver's licenses.}
    \label{fig:03-25}
\end{figure}

\item There is no guarantee that at least one specific feature to identify the document type in the new unseen types of ID documents that comes to the dataset can be identified.
\item The ID documents to be classified are the actual world samples affected by different lighting conditions when copied/photographed. There are several samples whose unique feature has been affected by the spotlight; thus, classifying the images based on just one unique feature can be affected by the explained reason. Figure \ref{fig:03-26} shows a sample in which the top banner keypoints of the document can not be extracted by the SIFT model due to the light spot.
\end{itemize}

\subsubsection{Methods Description}
Considering the above mentioned reasons, it has been chosen to detect more than one unique type identification feature in each document. By analyzing the dataset, it has been discovered that more than one text feature exists in each document that can identify the type of the document. As such,  considering those texts, the type of an ID document can be determined. Thus, to detect the text from the image, it has been decided to apply OCR to extract the characters used in each ID document.

\textbf{OCR}
A machine-driven translation method named optical character recognition (OCR) has been employed to identify text embedded in an image. In other words, OCR is a method that converts scanned or typed text files and handwritten typescript to editable typescript for subsequent handing, whether the images come from a scanned paper, a picture of an article, etc. OCR utilizes a visual mechanism to detect handwriting and document characters. This method is intended to process images containing text with minimal intervention from non-text results. Although the consistency of the input document affects the OCR output \cite{satyawan2019citizen}.

This method is extensively used as a method of data entrance from written paper data accounts, such as passport IDs, statements, bank statements, automatic receipts, commercial cards, mail, printouts of stationary information, or some other appropriate certification. It is a popular way of digitalizing reproduced typescripts such that they can be automatically modified, investigated, compressed, shown online, and used in machine procedures. This expertise enables computers to automatically read text. There are several uses for OCR, including identifying license plates, text extraction from natural scene photographs, and text extraction from scanned documents \cite{patel2012optical}. Today, the industry offers several OCR applications, including desktop OCR, server OCR, and web OCR. OCR accuracy may be influenced by the text pre-processing and segmentation algorithms used. Occasionally, text cannot be extracted from a picture due to the image's varying format, style, orientation, or complicated context. Any OCR tool's accuracy score ranges between 71\% and 98\% \cite{patel2012optical}.

In the current study, a metadata set has been developed, in which, all unique text data that could identify the type of each ID document has been manually detected. There are three kinds of texts or metadata in this project's library for each class.

\begin{itemize}
\item The text features that are unique to a specific type of document and cannot be found in any other documents.
\item The text features that are mutual in different types of documents, especially in documents that belong to the same state or territory. These texts usually identify the document type, country, state, or territory of the ID document. (e.g., "Driver Licence", "New South Wales").
\item The text features that are mutual in different types of documents that belong to a specific state or territory. These texts usually identify the sub-class of a specific document type (e.g., "Heavy Vehicle").
\end{itemize}

Table \ref{tab:03-01} shows the metadata-set for 4 different classes.

\begin{table}[h!]
\centering
\caption{Metadata samples for Australian driver's licenses.}
\label{tab:03-01}
\resizebox{\textwidth}{!}{%
\begin{tabular}{|l|l|l|l|l|l|l|l|}
\hline
\rowcolor[HTML]{C0C0C0}
\#         & \multicolumn{7}{c|}{\cellcolor[HTML]{C0C0C0}\textbf{Unique Text Features}}     \\ \hline
\textbf{1} & HEAVY    & VEHICLE & DRIVER   & LICENCE   & AUSTRALIAN & CAPITAL   & TERRITORY \\ \hline
\rowcolor[HTML]{EFEFEF}
\textbf{2} & Driver   & Licence & New      & South     & Wales      & Australia & -         \\ \hline
\textbf{3} & DRIVER'S & LICENCE & SOUTH    & AUSTRALIA & PLEASE     & CARRY     & -         \\ \hline
\rowcolor[HTML]{EFEFEF}
\textbf{4} & LEARNER  & PERMIT  & VICTORIA & AUSTRALIA & vicroads   & -         & -         \\ \hline
\end{tabular}%
}
\end{table}

OCR has been applied to extract all characters and then the characters in which there was no space have been accepted as a word. After that, all words extracted from the OCR have been compared to the metadata set. Then to classify the images based on the metadata set, a confidence level for each class has been assigned. Then the number of text features that matched each class's metadata have been assigned and divided that number to the total number of each class pre-defined metadata. Having two different classifiers confidence levels (SIFT and OCR), the fusion confidence level using the same fuzzy mean technique mentioned in approach 3 section calculated, and 100\% accuracy on 636 data has been achieved.

\subsubsection{Conclusion}

In this section, it has been explained why classification based on just one unique visual or text feature in a document is not leading us to the highest possible classification accuracy. To overcome the shortcomings of the feature detection-based classification in SIFT model, it was decided to use OCR to extract the unique text features of each ID document to identify the type of the ID document. Hence, metadata set based on the document type identifiable text features has been created. The extracted texts from OCR have been compared with each class's metadata. Then the percentage of the metadata of each class that has been detected in the sample document has been calculated. The calculated percentage has been considered and accepted as the confidence level of belonging of the sample document to a specific class. Then a fusion model has been created based on the mentioned confidence level of the OCR model and the confidence level of the previously explained SIFT-based classifier. Fuzzy mean logic has been applied to get the overall target class. Testing the fusion model on 636 data 100\% accuracy has been achieved.


\section{Experiments, Results, and Evaluation}

In this section, more details on the final proposed approach, which is believed to return the best results, have been presented. First, the used dataset and the pre-processing, which has been applied to get better and faster results on the final proposed method, are introduced, then the implementation specifications are explained. System development, testing, and performance analysis are other topics covered in the current section. Also, the software, programming languages, libraries, and functions were used to implement each model have been described. Finally, all four methods' results were compared and analyzed, and the final proposed fusion model, which results in the highest accuracy, has been described. Then the proposed fusion model results have been compared with the results of similar research in the field of identity document classification, which has been studied and discussed in the literature review section.

\subsection{Dataset}

The current research is conducted in collaboration with LOCII/TRUUTH holdings. Truuth is a subsidiary of locii Innovation Pty Ltd, a privately held Australian company in Sydney in 2020, targeted to solve the fake identity problem for consumers and businesses. They operate in the field of Computer Software, Cybersecurity, Blockchain, and Identity Verification. Their objective is to build a portfolio of products that users and businesses could use throughout their journey of any online services, including but not limited to accessing a business site, verification of identity for a bank account, visa, passport driving license, etc. They are trying to achieve their goals by addressing the gaps in the current KYC (Know Yout Customer) processes and considering several biometrics parameters and liveness testing for a smooth customer experience with a high level of confidence in legitimate identity. In the process, Truuth is attempting to address this need by working on multiple products in their platform, including but not limited to Truuth KYC, Identity, Bio-pass, Liveness, etc., and developing multiple desktop and mobile applications. Their products collectively address the fake identity problem in the digital and online market toward solid ownership of their customers' identities.

The first phase of Truuth ID document classification project, which has been explained in this research, was to classify the Australian Identity Documents, mainly Australian driver's licenses and Medicare cards. The dataset which has been applied in this study includes data from internet resources, friends and family documents, and documents from Truuth customers. The data which has been collected from internet resources are both fake or genuine documents with various resolutions, whereas the data from Truuth customers and friends and family are genuine, original, and valid (not expired); in addition, the quality and resolution of the second-mentioned type of the data are generally much higher than publicly available data.

Apart from usual national ID documents, such as passport and birth certificates, there are different types of driver's licenses in Australia. This country consists of six states and two territories\footnote{https://www.australia.gov.au/statess}, in which there are different types of driver's licenses such as full driver's license, learner driver's license, provisional driver's license, heavy vehicle driver's license, etc. The visual features and logo of the driver's licenses of a specific state or territory are almost similar; however, there are minor differences in some of their visual and text features. The main difference between the driver's licenses within a particular state/territory is the document's color. Samples from 25 different driver's licenses and Australian resident Medicare cards were accessible to this project. The total number of data was 636, including 608 driver's licenses and 28 Medicare cards. The number of samples varied in different classes. The highest number of samples belonged to NSW full driver's license with 120 samples, and the smallest samples belonged to ACT driver's licenses with eight samples.

\subsection{Pre-Processing}
The current project is supposed to be used in real-time or real-world applications and digital identity services; hence, the process must be designed so that when a sample data comes in, the classification process to work independently of any manual pre-processing i.e., cropping and rotation adjustment. For this reason, no manual pre-processing on the test set has been applied. However, test samples were resized, and their resolution has been decreased to minimize the computation time and eliminate the noises that could have been detected as visual features. In other words, with resolution reduction,  the critical features of the image can be highlighted. By try and error method, it has been decided that the resolution of 425*270 pixels on the data set returned the best results. As mentioned in the dataset section, the train set images are cropped precisely, and the rotation has been adjusted for all source images.

\subsection{SIFT + OCR, ID Document Classifier}
The implementation has been done so that when a new class comes in, there was no need to change the whole implementation, and just by modifying the train set,  the samples of the new target class could be classified. In other words, the model has been implemented in such a way that it could be developed for any ID document. All techniques and formulas have been implemented in existing programming functions and components in Python. In this section, first, each method's implementation and preliminary results have been described, then how to combine the models to create the fusion model has been explained, and then the results of the fusion model have been illustrated.

\subsubsection{Classification Based on the SIFT}
The SIFT model patent for industrial use was expired in 2020 after 20 years. OpenCV (Open-Source Computer Vision), developed by Intel, is a real-time machine vision library that includes the SIFT model\footnote{https://opencv.org}. The OpenCV functions in Python to implement the visual feature-based classification using the SIFT model have been used in this project. After applying the SIFT feature extractor on the sample image and all source/train images to extract the key-points, a model to match the key-points of sample key-points with the source image key-points, has to be applied. In order to create the matches, there are two methods available which are also implemented in OpenCV:

\textbf{Brute-Force matcher} which using a specific distance calculation matches the descriptor of one feature with all descriptors of all features of the source or training images. According to the description, Brute-Force method is obviously time-consuming.

\textbf{FLANN (Fast Library for Approximate Nearest Neighbors) matcher} in which a number of optimized algorithms are used for fast nearest neighbor search. Therefore, because FLANN matcher works faster than Brute-Force matcher, it is usually used in large datasets.

\begin{figure}[h!]
    \centering
    \includegraphics[width=0.62\textwidth]{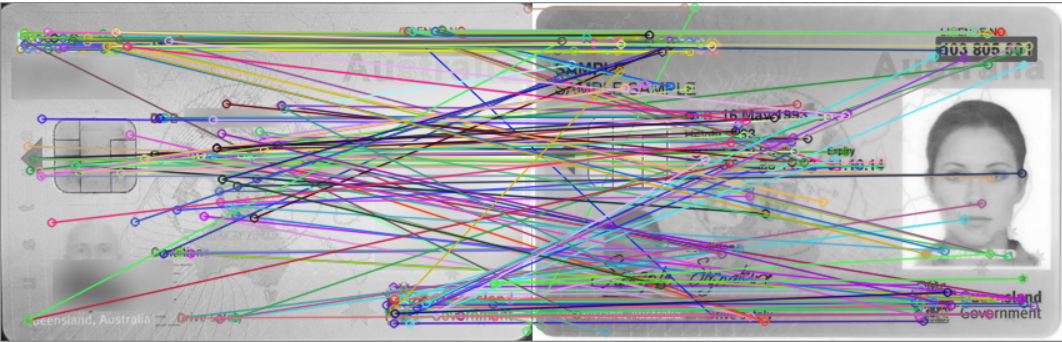}
    \caption{Matching features of two images.}
    \label{fig:04-27}
\end{figure}

The goal of the current research was to get the highest possible accuracy in ID document classification. Therefore both Brute-Force matcher and FLANN matcher have been implemented, and Brute-Force matcher has been accepted as the project's feature matching method due to getting slightly higher accuracy. The accepted class has been considered the class of the source image with the highest matching features with the sample image as the target class. Figure \ref{fig:04-27} illustrates a matching result sample.

\subsubsection{classification Based on OCR:}
To implement an OCR model, the google cloud vision library in Python has been used. All letters and numbers with no space in between have been accepted as a single word. An excel file has been created in which each row represents a specific class. Each row contains multiple keywords which can identify the type of the relevant identity document. After extracting all words of an image using OCR, the extracted words have been compared to the unique words of each class. Finally, the class with the most matched words has been accepted as the target class. The confidence level is the percentage of the matched words with each class.

\subsubsection{Confidence Level for the SIFT-Based Classification}
As mentioned in the methodology section, the feature matching-based classification results are based on the numbers which indicate the amount of matched features between each sample image with any other images. Also, it has been described that the winner class is the class with the highest number of matched features. As a result, when an image is classified based on feature matching,  the confidence level in terms of the classification results can not be expressed. Therefore, a method in which the confidence score of each classification result can be expressed shall be developed. Consequently, a novel fusion model based on statistics and regression has been proposed.  In this model, the number of matched features of each classification result has been normalized using Z-score to illustrate the distribution of each number and the distance of the score with the mean value of the distribution. Thus, not only, the data will be normalized in a limited range to compare the results with each other, but also the Z-scores can be analyzed, and the behavior of a sample image and its similarities with the train set can be explained. To calculate the Z-score the Python Statistics library has been used.

After normalizing the results, the probability of occurrence of each data in the appropriate class has to be calculated. In this project's classification,  only the existence or non-existence of the sample image in the correct class has been considered, which means this was a binary classification problem. As explained in the methodology section, having a set of data (Z-score) that belong to a binary class,  a simple logistic regression can be applied to get the probability of the existence of each data in the target class. A logistic regression in Python using the "sklearn" library has been implemented. To train the logistic regression 636 samples have been classified, in which different types of samples with different resolutions and lighting conditions have been tested. For each sample, the number of matching features with all source images has been extracted, and the target class of each sample has been specified as class 1. Then the logistic regression to get the probability has been applied and the probability has been considered the confidence score. The regression has been trained with the images which have been classified correctly, as well as the samples which have been miss-classified. Then the pre-trained regression has been saved and used to calculate the confidence level of the other SIFT-based classification results. Table \ref{tab:04-01} shows the raw score, z-score, target class, and the confidence level of 3 samples.

\begin{table}[h!]
\centering
\caption{Raw score and z-score of 3 samples.}
\label{tab:04-01}
\begin{tabular}{|c|c|c|c|c|c|}
\hline
\rowcolor[HTML]{C0C0C0}
\multicolumn{2}{|c|}{\cellcolor[HTML]{C0C0C0}\textbf{Sample 1}} & \multicolumn{2}{c|}{\cellcolor[HTML]{C0C0C0}\textbf{Sample 2}} & \multicolumn{2}{c|}{\cellcolor[HTML]{C0C0C0}\textbf{Sample 3}} \\ \hline
\textbf{Raw Score}              & \textbf{Z Score}              & \textbf{Raw Score}              & \textbf{Z Score}             & \textbf{Raw Score}              & \textbf{Z Score}             \\ \hline
\rowcolor[HTML]{EFEFEF}
512                             & 3.51                          & 120                             & 3.6                          & 69                              & 4.56                         \\ \hline
354                             & 2.16                          & 49                              & 0.83                         & 28                              & 1.1                          \\ \hline
\rowcolor[HTML]{EFEFEF}
304                             & 1.73                          & 40                              & 0.48                         & 20                              & 0.43                         \\ \hline
202                             & 0.86                          & 37                              & 0.37                         & 16                              & 0.09                         \\ \hline
\rowcolor[HTML]{EFEFEF}
162                             & 0.52                          & 35                              & 0.29                         & 15                              & 0.01                         \\ \hline
154                             & 0.45                          & 33                              & 0.21                         & 14                              & -0.07                        \\ \hline
\rowcolor[HTML]{EFEFEF}
54                              & -0.39                         & 31                              & 0.13                         & 12                              & -0.23                        \\ \hline
52                              & -0.41                         & 31                              & 0.13                         & 11                              & -0.32                        \\ \hline
\rowcolor[HTML]{EFEFEF}
44                              & -0.48                         & 28                              & 0.021                        & 9                               & -0.49                        \\ \hline
44                              & -0.48                         & 22                              & -0.21                        & 9                               & -0.49                        \\ \hline
\rowcolor[HTML]{EFEFEF}
37                              & -0.54                         & 10                              & -0.68                        & 6                               & -0.74                        \\ \hline
\end{tabular}
\end{table}

The last step to combine the methods to develop a single fusion classifier was to calculate the mean value of the two classifier confidence scores, and then the class to which the highest confidence score belongs has been accepted as the target class. The first three highest confidence levels have been considered to feed the final classifier approach. Table \ref{tab:04-02} shows how the fusion model determines the correct class.

\begin{table}[h!]
\centering
\caption{How the fusion model determines the target class.}
\label{tab:04-02}
\begin{tabular}{|c|c|c|c|c|c|c|}
\hline
\cellcolor[HTML]{C0C0C0}                            & \multicolumn{2}{c|}{\cellcolor[HTML]{C0C0C0}\textbf{SIFT Classifier}} & \multicolumn{2}{c|}{\cellcolor[HTML]{C0C0C0}\textbf{OCR Classifier}} & \multicolumn{2}{c|}{\cellcolor[HTML]{C0C0C0}\textbf{Fusion Model}} \\ \cline{2-7}
\multirow{-2}{*}{\cellcolor[HTML]{C0C0C0}Sample} & Detected Class                  & Probability                         & Detected Class                  & Probability                        & Mean                       & Target Class                    \\ \hline
                                                    & \cellcolor[HTML]{EFEFEF}1       & \cellcolor[HTML]{EFEFEF}0.99875     & \cellcolor[HTML]{EFEFEF}1       & \cellcolor[HTML]{EFEFEF}0.9768     & \cellcolor[HTML]{EFEFEF}0.987    &                                 \\ \cline{2-6}
                                                    & 6                               & 0.00062                             & 6                               & 0.01381                            & 0.007                            &                                 \\ \cline{2-6}
\multirow{-3}{*}{1}                                 & 5                               & 0.00062                             & 4                               & NA                                 & NA                               & \multirow{-3}{*}{\textbf{1}}    \\ \hline
                                                    & 8                               & 0.994                               & 26                              & 0.01                               & NA                               &                                 \\ \cline{2-6}
                                                    & \cellcolor[HTML]{EFEFEF}15      & \cellcolor[HTML]{EFEFEF}0.031       & \cellcolor[HTML]{EFEFEF}15      & \cellcolor[HTML]{EFEFEF}0.98       & \cellcolor[HTML]{EFEFEF}0.505    &                                 \\ \cline{2-6}
\multirow{-3}{*}{2}                                 & 9                               & 0.001                               & 25                              & 0.01                               & NA                               & \multirow{-3}{*}{\textbf{15}}   \\ \hline
\end{tabular}
\end{table}

\subsection{Results and Evaluation}
For evaluation of the system and the results, Receiver Operating Characteristic (ROC) curve analysis has been used because ROC curve analysis is one of the most popular methods of model evaluation. The ROC curve is a measure of efficiency in classification problems. This curve is widely used in signal recognition theory, image processing systems in radiological applications, and in various fields of medicine such as cognitive testing and treatment regimens.
A graphical demonstration of the degree of sensitivity or true prediction versus false prediction is known as ROC Curve. It is applicable where the separation threshold varies in a binary system. The ROC is also represented by plotting the right-predicted positives (TP) versus the wrong-predicted positives (FP). The number generated by calculating the area under the ROC curve represents one aspect of performance. The Area Under Curve, or AUC, can be between zero and one. One means an excellent prediction, and an output of 0.5 means a random prediction. Assuming a two-category (binary) prediction problem where the outputs are shown as positive (P) or negative (N), the confusion matrix can be represented. By sorting the categories of the classification problem, this matrix demonstrates the performance of the classification algorithm based on the input data set. The confusion matrix returns True Negative (TN), True Positive (TP), False Positive (FP), and False Negative (FN).

ROC curve analysis uses different criteria to measure system performance, including sensitivity and specificity, which are described below:
(i)~\textbf{sensitivity} indicates the correctly predicted values against all positive outputs. This value is also called the True Positive Rate (TPR); and
(ii)~\textbf{specificity} indicates negatively predicted values versus all negative outputs.
The ROC chart is based on TPR and FPR, which are the same as false-positive rates. The FPR value can be obtained by subtracting the specificity value from one. This value is equivalent to the ratio of the number of false positives to the total number of negatives.

\begin{figure}[h!]
    \centering
    \includegraphics[width=0.9\textwidth]{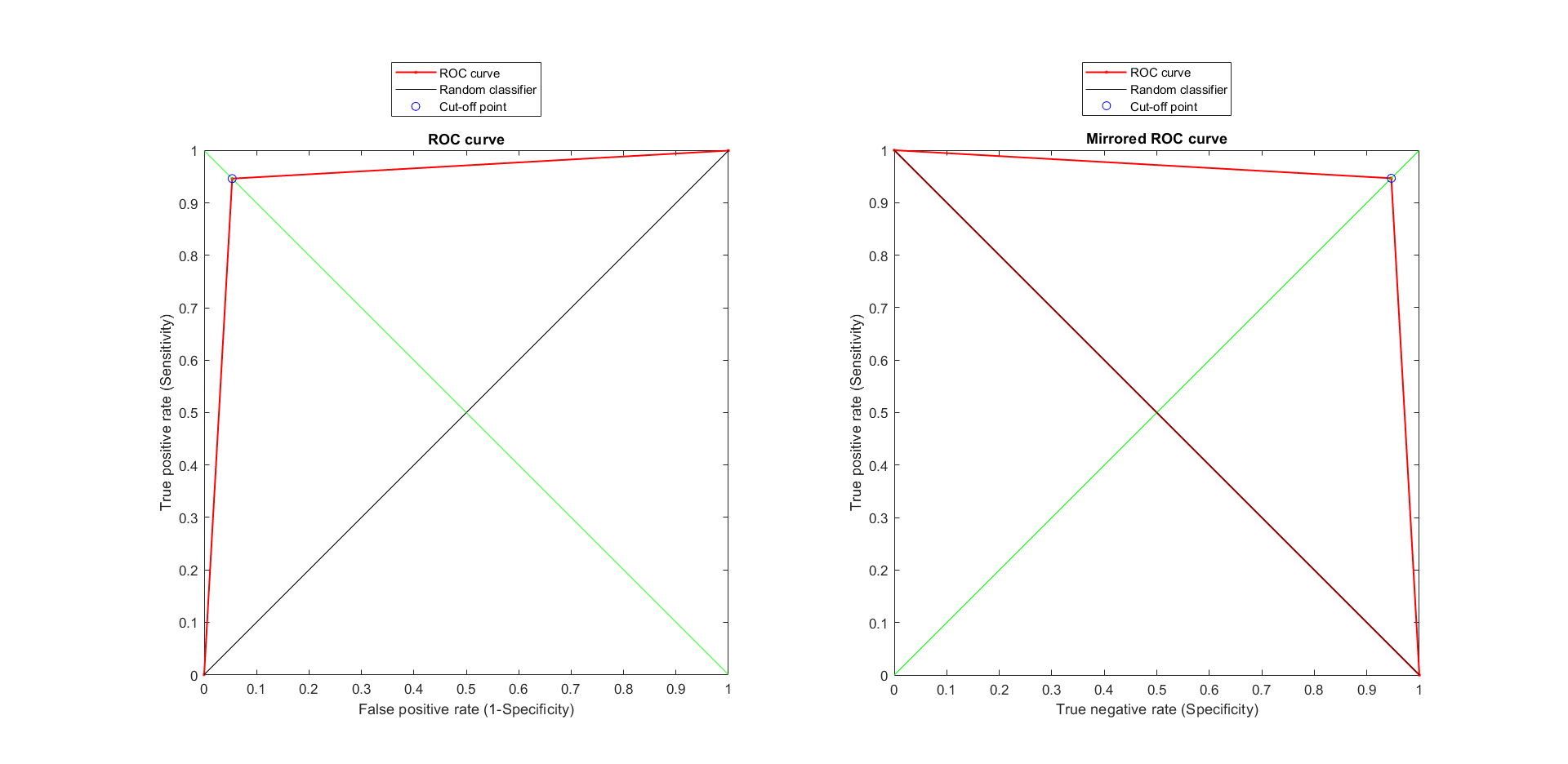}
    \caption{ROC curve for the SIFT classification.}
    \label{fig:04-ROC-SIFT}
\end{figure}

The ROC curve analysis has been used to illustrate the results. The results of each method (SIFT and OCR), as well as the proposed fusion model, have been analyzed separately. Each sample returns 27 results, which equals the number of classes. The target class represents as class 1, and the rest of the classes indicate as class 0. All zero or negative classes have been considered as one unique class to get more precise results. The reason is that detecting the class in which the model miss-classifies the sample is not essential or informative. What matters is to be able to detect the correct class of each ID document. Figure \ref{fig:04-ROC-SIFT} shows the ROC curve applied on the results of testing the SIFT-based classification on 636 data. We get 24 miss-classifications which leads us to 96.22\% accuracy.

\begin{figure}[h!]
    \centering
    \includegraphics[width=0.9\textwidth]{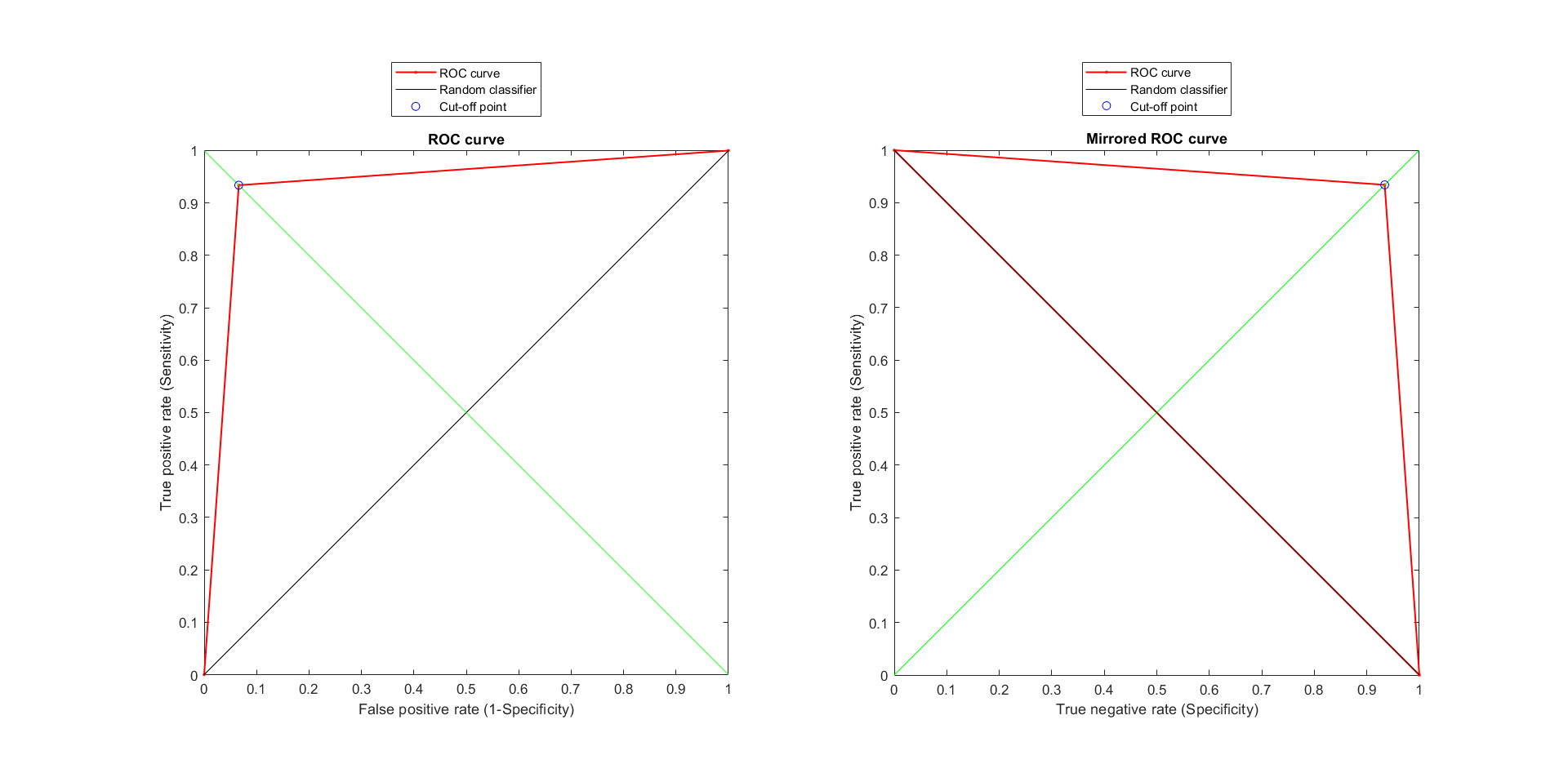}
    \caption{ROC curve for the OCR classification.}
    \label{fig:04-ROC-ocr}
\end{figure}

The ROC diagram for the test data set, classified based on the OCR is shown in Figure \ref{fig:04-ROC-ocr}. According to this chart, the true positive rate (TPR) is equal to 0.94 and the area under curve is 0.9747.

\begin{figure}[h!]
    \centering
    \includegraphics[width=0.9\textwidth]{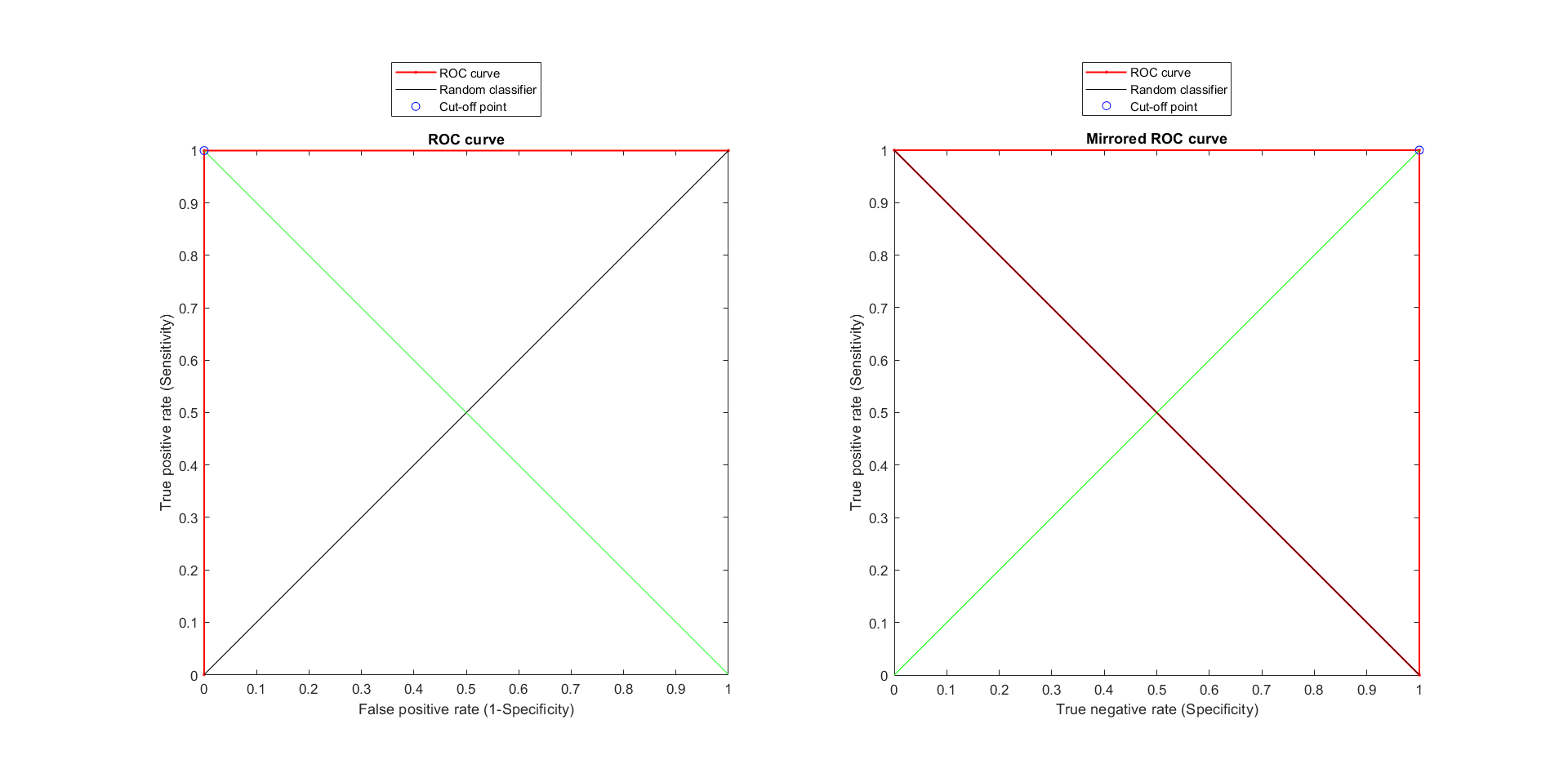}
    \caption{ROC curve for the SIFT + OCR, fusion classification.}
    \label{fig:04-ROC-fusion}
\end{figure}

The ROC curve, for the fusion model, tested on 636 data is illustrated in Figure \ref{fig:04-ROC-fusion}.  100\% accurac has been acheived, thus obviously according to the ROC, 1 for the true positive rate and 0 for the false positive rate has been reached. Also, the area under the curve is equal to 1.

Tables \ref{tab:04-03} and \ref{tab:04-04} represent the confusion matrix, in addition to the information such as sensitivity, specificity, and accuracy of the SIFT and OCR-based classification, as well as the proposed fusion model, obtained from ROC analysis which is used to analyze the performance of the algorithm. The analysis is a binary analysis in which the classification result is either true (1) or false (0), therefore TP and TN values are equal, similarly, FP and FN values are equal.

\begin{table}[h!]
\centering
\caption{Confusion matrix.}
\label{tab:04-03}
\begin{tabular}{|c|c|c|c|c|}
\hline
\rowcolor[HTML]{C0C0C0}
\textbf{Classifier} & \textbf{True Positive} & \textbf{False Positive} & \textbf{True Negative} & \textbf{False Negative} \\ \hline
\textbf{SIFT}       & 612                    & 24                      & 612                    & 24                      \\ \hline
\rowcolor[HTML]{EFEFEF}
\textbf{OCR}        & 602                    & 34                      & 602                    & 34                      \\ \hline
\textbf{SIFT + OCR} & 636                    & 0                       & 636                    & 0                       \\ \hline
\end{tabular}
\end{table}

\begin{table}[h!]
\centering
\caption{ROC analysis.}
\label{tab:04-04}
\begin{tabular}{|c|c|c|c|c|}
\hline
\rowcolor[HTML]{C0C0C0}
\textbf{Classifier} & \textbf{AUC} & \textbf{Accuracy} & \textbf{Sensitivity} & \textbf{Specificity} \\ \hline
\textbf{SIFT}       & 0.9818       & 96.22\%           & 0.9623               & 0.9623               \\ \hline
\rowcolor[HTML]{EFEFEF}
\textbf{OCR}        & 0.9747       & 94.65\%           & 0.9465               & 0.9465               \\ \hline
\textbf{SIFT + OCR} & 1            & 100\%             & 1                    & 1                    \\ \hline
\end{tabular}
\end{table}

\subsection{Discussion}
In this section,  the implementation details of the proposed fusion model have been elaborated. The results of the fusion model, as well as analysis of the results of each section of the fusion model individually, have been reviewed and analyzed. The goal of the current study was to find the best possible approach to identify the type of different identity documents. A novel approach to generate the confidence level for feature matching and similarity-based classifications resulted in a novel fusion model for highly accurate image classification. Furthermore, the proposed method enabled the project to calculate the confidence score of the classification model; therefore, according to the different situations and projects and based on the sensitivity and importance of the classification result, it can be decided which results should be accepted and which results are not reliable.

Moreover, considering the scenarios in which the result of the classification is crucial, such as the motivation of the project's sponsor, Truuth,  based on the reliable confidence level, another automated step can be applied to the model to minimize the risk of miss-classification, this will be discussed in the future work section.


\section{Conclusion}

The unauthorized use of an individual's personal information to commit a crime or deceive the individual or a third party is known as identity fraud. The majority of identity frauds have been done for financial advantages, forging identity documents for criminal acts, or supporting applications for governmental agencies. Currently, most of the document authentication services are manually performed, which are error-prone due to the nature of the human-based process, also time-consuming. Furthermore, to perform a diligent assessment, the original version of the document should be provided, which is a problem on its own.  To achieve a high level of user experience and to streamline the process of document authentication, to spot the fraudulent or falsified documents based on a photo or image of the document, in the presented research, a novel document classification step has been proposed that can be added before authentication of identity documents.

In this study, after explaining the main research problem, the importance and necessity of research and objectives have been investigated. Then, in section 2, research and materials related to the subject of the present project have been reviewed. Also, the previous literature has been studied, having said that, not significant research material has been available targeting the same problem. In section 3, different methods that have been applied step by step have been described; also, the analysis method of the results based on the dataset has been discussed, and the attempt to improve the model by applying an appropriate method has been explained. In the same section,  the implementation and the accuracy of all approaches have been briefly explained. section 4 contains the implementation, results, and evaluation of the final approach, which resulted in the highest accuracy. The results of the fusion model and its individual sections, based on the OCR curve analysis method, have been reported and analyzed. And, the results and performance of the novel proposed model has analyzed and has been compared with the results and reported accuracy of previous studies, which were mentioned in the literature review section

\subsection{Summary and Discussion}

The main goal of this study was to achieve the most accurate approach to classify the ID documents.

It has been concluded that the current studies and approaches on image classification could not direct the project to satisfy its' goal. Hence, considering the studies conducted in this field, the preliminary model has been conducted by replicating one of the studies \cite{simon2015fine} in which the dataset condition and characteristics were very similar to the dataset in this research; moreover, the results reported in the mentioned research were considerable. According to the final results of each approach, The analysis of the research dataset, and the strengths and weaknesses of the existing methods, the novel proposed model has improved the outcomes in 4 steps which have been explained in sections 3 and 4 in detail.  As discussed,  ID documents are highly confidential data, so gathering a large dataset to train robust machine learning models have been challenging and a constraint for the research project; thus it has been decided to take an approach that works on matching samples (test data) with a source document, and the classification has been done based on the similarity of the sample image with the source images. To achieve the goal,  a novel approach has been proposed that enabled the model to return a confidence level for each classification based on the number of matches in the SIFT model, which can be used in any model that works based on the rate of the similarity of each sample with each source data. Table \ref{tab:05-01} shows the results of the approach presented in this study. The proposed approach has been tested on the accessible data set at the time, and as more data has been collected, the new models have been tested on the larger dataset.

\begin{table}[h!]
\centering
\caption{The comparison of the current research proposed approaches.}
\label{tab:05-01}
\begin{tabular}{|l|c|c|}
\hline
\rowcolor[HTML]{C0C0C0}
\textbf{Approach}                   & \textbf{Number  of Test Data} & \textbf{Accuracy} \\ \hline
\textbf{HOG + Color Name + SP3}     & 105                           & 63.8\%            \\ \hline
\rowcolor[HTML]{EFEFEF}
\textbf{HOG + Fuzzy Color Name}     & 105                           & 72.38\%           \\ \hline
\textbf{SIFT + Feature Engineering} & 152                           & 98.02\%           \\ \hline
\rowcolor[HTML]{EFEFEF}
\textbf{SIFT + OCR}                 & 636                           & \textbf{100\%}    \\ \hline
\end{tabular}
\end{table}

\subsection{Comparison with the
Previous Studies}
As mentioned in section 2, there are not many studies available that have worked on the same problem. The reason is that although enormous image classification studies have been published so far, just a few of them conducted on the identity document classification specifically. The table \ref{tab:05-02} shows a summary of some of the researches on ID document classification which has been mentioned in the literature review section. According to the results demonstrated in table \ref{tab:05-02}, the novel approach proposed in this thesis is the most accurate approach with 100\% accuracy among the previous ID document classification approaches.

\begin{table}[h!]
\centering
\caption{The comparison of the current research with the previous studies.}
\label{tab:05-02}
\resizebox{\textwidth}{!}{%
\begin{tabular}{|l|l|l|}
\hline
\rowcolor[HTML]{C0C0C0}
\textbf{Main Methods}           & \textbf{Dataset Size}                   & \textbf{Accuracy}                                                                   \\ \hline
HOG + Color-name + SP3 \cite{simon2015fine}          & 375                                     & 97.7\%                                                                              \\ \hline
\rowcolor[HTML]{EFEFEF}
CNN + Feature Extraction \cite{awal2017complex}        & 3042                                    & 96.6\%                                                                              \\ \hline
CNN \cite{vilas2018classification}                             & 144                                     & 98\%                                                                                \\ \hline
\rowcolor[HTML]{EFEFEF}
CNN \cite{sicre2017identity}                             & Tested different datasets including FRA & \begin{tabular}[c]{@{}l@{}}Different results reported\\(60\% - 98.2\%)\end{tabular}                                                          \\ \hline
SURF \cite{almaksour2016classification}                            & 2494                                    & 95.75\%                                                                             \\ \hline
\rowcolor[HTML]{EFEFEF}
RANSAC \cite{skoryukina2019fast}                          & MIDV-500                                & \begin{tabular}[c]{@{}l@{}}Different results reported\\(41.7 \% - 76\%)\end{tabular}                                                          \\ \hline
OCR \cite{satyawan2019citizen}                             & Different datasets                      & \begin{tabular}[c]{@{}l@{}}Different results reported\\ (60\% - 98.2\%)\end{tabular} \\ \hline
\rowcolor[HTML]{EFEFEF}
\textbf{Current Proposed Model} & \textbf{636}                            & \textbf{100\%}                                                                      \\ \hline
\end{tabular}%
}
\end{table}

\subsection{Future Work}
As discussed in this study, the outcome of this research will feed another phase to authenticate the identity documents based on the defined type of the document using the currently proposed method. Therefore, the goal of the current study was to achieve the most accurate ID document classification results to mitigate the probability of any failure or error in ID document authentication. Hence, not only to eliminate all possible classification failure but also to be able to automatically add the new documents to the training set, to be able to classify new document worldwide,  another approach to the classification results will be added to enhance the accuracy and to make the classifier robust in terms of facing unseen samples.
As mentioned in the literature review section, crowdsourcing is one of the common approaches used to enhance the accuracy of the classification results. Crowdsourcing is the process of achieving a shared goal by gathering collective knowledge of a group of ordinary or expert individuals. This process naturally increases the likelihood of solving a complex problem innovatively, enabling the business to reach a broader range of resources/labor. Filling the knowledge gap is one of the many benefits of crowdsourcing. It also helps businesses accelerate while lowering operational costs due to using more resources with lower or no overhead cost; thus, businesses will become more scalable. We plan to leverage rule-based~ technologies~\cite{DBLP:conf/icse/TabebordbarB18,DBLP:conf/wise/TabebordbarBBB19,DBLP:journals/dase/TabebordbarBBB20} and storytelling techniques~\cite{DBLP:conf/www/BeheshtiTB20,DBLP:conf/wise/TabebordbarBB19,DBLP:journals/access/GhodratnamaBZS20} to turn the results into arguments and insights.

In the future steps, it is suggested to leverage the crowdsourcing technique to take samples from clusters/classes, share them with crowd workers, and use the crowd's knowledge to personalize the results using cognitive recommendations~\cite{DBLP:journals/algorithms/BeheshtiYMGGE20,DBLP:conf/pacis/YakhchiBGO19,DBLP:conf/wise/YakhchiGB18,DBLP:journals/access/YakhchiBGO020}.
%
Furthermore, it is suggested that a feedback loop be presented to continuously take samples from the clusters and refine the categories of classified documents. All mentioned will be applied to the samples in which the confidence level of the proposed fusion classifier is less than a pre-defined threshold.


\section*{Acknowledgements}
- I acknowledge the AI-enabled Processes (AIP) Research Centre at Macquarie University and Locii Holdings Pty Ltd (trading as Truuth) for funding my research scholarship.

\bibliographystyle{abbrv}
\bibliography{ms}

\end{document}